\documentclass{article}

\usepackage{arxiv}

\usepackage[utf8]{inputenc} % allow utf-8 input
\usepackage[T1]{fontenc}    % use 8-bit T1 fonts
\usepackage{hyperref}       % hyperlinks
\usepackage{url}            % simple URL typesetting
\usepackage{booktabs}       % professional-quality tables
\usepackage{amsfonts}       % blackboard math symbols
\usepackage{nicefrac}       % compact symbols for 1/2, etc.
\usepackage{microtype}      % microtypography
\usepackage{lipsum}
\usepackage[title]{appendix}

\usepackage{xcolor}

\usepackage[linesnumbered,ruled,vlined]{algorithm2e}

\usepackage{stmaryrd}
\usepackage{empheq}
\usepackage{wrapfig}
\usepackage{multicol}
\usepackage[english]{babel}
\usepackage{blindtext}
\usepackage{tabularx}
\usepackage{pdflscape}
\usepackage{multirow}
\usepackage[inline]{enumitem}
\usepackage{subcaption}

\usepackage{floatrow}
\newfloatcommand{capbtabbox}{table}[][\FBwidth]

\newcommand{\eqname}[1]{\tag*{#1}}% Tag equation with name

\SetCommentSty{mycommfont}

\SetKwInput{KwInput}{Input}
\SetKwInput{KwOutput}{Output}
\SetKwInput{KwInit}{Initialization}
\SetKwProg{Fn}{Function}{}{}
\SetKwInput{TM}{Modelisation Procedure}

\title{Unsupervised clustering of series using dynamic programming}

\author{
  Karthigan Sinnathamby\thanks{Work done while in Master Data Science at EPFL (Lausanne, Switzerland)}\\
  Visium\\
  Lausanne, Switzerland\\
  \texttt{karthigan.sinnathamby@visium.ch} \\
   \And
    Chang-Yu Hou \\
  Schlumberger-Doll Research \\
  Cambridge, MA\\
  \texttt{CHou2@slb.com} \\
     \And
 Lalitha Venkataramanan \\
  Schlumberger-Doll Research \\
  Cambridge, MA\\
  \texttt{LVenkataramanan@slb.com} \\
       \And
 Vasileios-Marios Gkortsas \\
  Schlumberger-Doll Research \\
  Cambridge, MA\\
  \texttt{VGkortsas@slb.com} \\
\And
François Fleuret \\
  University of Geneva\\
  Geneva, Switzerland\\
  \texttt{francois.fleuret@unige.ch} \\
}

\begin{document}
\maketitle

\begin{abstract}
We are interested in clustering parts of a given single multi-variate series in an unsupervised manner. We would like to segment and cluster the series such that the resulting blocks present in each cluster are coherent with respect to a known model (e.g. physics model). Data points are said to be coherent if they can be described using this model with the same parameters. We have designed an algorithm based on dynamic programming with constraints on the number of clusters, the number of transitions as well as the minimal size of a block such that the clusters are coherent with this process. We present an use-case: clustering of petrophysical series using the Waxman-Smits equation.\end{abstract}

% keywords can be removed
\keywords{Unsupervised Clustering \and Series \and Dynamic Programming}

\section{Introduction}
Unsupervised clustering is a branch of machine learning that aims to categorize the data based on the self-similarity. In other word, data-points in the same group (called a cluster) are more similar to each other than to those in other groups. This task can be achieved by various algorithms (the well-known K-means or spectral clustering but also hierarchical clustering \cite{7100308} or density-based clustering \cite{malzer2019hybrid}) that differ significantly in their understanding of what constitutes a cluster and how to efficiently find them.

In many cases, there exist models/functions, governed by a finite set of parameters, providing either physics or phenomenology correlations between input data. The presence of these models can in principle be used to characterize clusters (cluster characterization) because one can define a loss function to measure how well a point belongs to this cluster (cluster affiliation). Hence, given a multi-variate series, one can cluster it such that the resulting blocks present in each cluster are coherent with respect to prescribed physics correlations. In other words, data points are clustered together if they can be described using the assumed model with the same modeling parameters.

Furthermore, we are interested in clustering multi-variate series which have temporal/spatial correlation such that the sequential data points have higher likelihood to belong to the same cluster. Namely, we expect that the cluster assignment should present a sparse clustering and consequently well defined blocks of points. However, the generic unsupervised clustering scheme often does not take this temporal/spatial correlation into account, which can overfit the data and lead to the too fragmented clustering. Therefore, it is of interests to establish a clustering algorithm that respect to the sequential correlation of the data.

In this work, we propose an unsupervised clustering algorithm that suits the following objective: (1) The clustering criterion is based on the physics coherence. (2) The sequential correlation of data series is properly respected. The general algorithm is inspired from the well-know K-means algorithm, which is a special case of the proposed algorithm where the cluster characterization is merely the mean point of the cluster and the cluster affiliation the L2 loss with this mean point. In order to properly taking into account the temporal/spatial correlation, the proposed algorithm will constraint the number of cluster, the number of transition (number of times we change to a different cluster), and the minimal size (number of data points) of any block in the final assignment. To have an efficient clustering algorithm, we will derive a recurrence relation automatically satisfying all the required constraints, and make use of dynamic programming to correctly find the optimal assignment.

%On the one hand, the algorithm will be independent of the two assumptions, i.e. it presents a general framework for any cluster characterization/affiliation pair. On the other hand, as we are interested in series which yield temporal/spatial correlation, the cluster assignment should present a sparse clustering (present well defined blocks of points) and not a dense one which will overfit the data (too fragmented). Therefore, the proposed algorithm will constraint the number of cluster, the number of transition  (number of times we change to a different cluster) as well as the minimal size of any block of points in the final assignment.

%The general algorithm is inspired from the well-know K-means algorithm, which is a special case of the proposed algorithm where the cluster characterization is merely the mean point of the cluster and the cluster affiliation the L2 loss with this mean point. In order to put those constraints, we will make use of dynamic programming to efficiently and correctly find the optimal assignment.

\section{Problem Definition}

We consider a multi-variate series $X$ and we would like to cluster this series among $C$ clusters. We suppose that each point of the series is of dimension d, in other words $X \in \mathbb{R}^{T \times d}$ where $T$ is the total number of points. We note $x_{t} \in \mathbb{R}^{d}$ the vector data at step $t$ and $y_{t} \in \{1,...C\} $ the cluster assignment for the point $x_{t}$. Moreover, we denote $Y$ the vector $[y_{1},...y_{T}]$.

We consider a modelisation function $M_{char}(X,Y,C)$ shortly noted $M_{char}$ and a cost function of the form $ f(x,w): \mathbb{R}^{d} \rightarrow  \mathbb{R}^{+} $. Given a clustering assignment $Y$, $M_{char}$ computes for each cluster $c$ (among the $C$ clusters) some weights $w_{c} \in \mathbb{R}^{d_{w}}$ where $d_{w}$ is the dimension of the weights, obtained through a training procedure over the points in $X$ assigned to the cluster $c$ (cluster characterization). We note $W \in \mathbb{R}^{C \times d_{w}}$ the matrix where the rows consist of each vector $w_{c}$. $f(x,w_{c})$ is the cost evaluating how bad the single point $x$ is fit to the cluster $c$ (cluster affiliation) given its weight $w_c$.

The goal is to find the best assignment $Y$ that minimizes the following objective:

$$ 
L_{f}(X,Y,W) = \sum_{c=0}^{C-1} \sum_{\substack{0 \leq t \leq T-1{}\\y_{t}=c}} f(x_{t},w_{c})
$$

We detail what is $M_{char}(X,Y,C)$ and $f(x,w_{c})$ in the case of K-means as an example in algorithm \ref{k_means}. The objective function in the K-means is the pairwise L2 distance between the points.

\begin{algorithm}
\DontPrintSemicolon
$f(x,w) = ||x - w||^{2}$
\BlankLine
\Fn{$M_{kmeans,char}$(X,Y,$C$)}{
	W = empty array of size $C$
	\BlankLine
	\For{$c\gets1$ \KwTo $C$} { 
		tmp = $\{ X[t] \mid Y[t]=c\}$
		
		\If {len(tmp) $\geq$ 1} {
			W[c] = Mean(tmp)
		}
		\Else {
		Remove cluster c
		}
		
	}

	return W
}

\caption{Cluster characterization and affiliation for $k_{means}$}
\label{k_means}
\end{algorithm}

\section{Algorithm Definition}

Given any $M_{char}(X,Y,C)$ and $f(x,w_{c})$, we propose the clustering algorithm in algorithm \ref{algo3} to minimize the given objective. The parameter $\epsilon$ defines the maximum difference between two cost values to consider that the cost has converged (to be chosen depending on the cost scale and the desired precision).

The idea is a generic version of the classical K-means algorithm. We start from a random assignment, we compute the weights of the clusters (cluster characterization) using the $M_{char}$. Then, for each point we assign the cluster that gives the lowest value in cluster affiliation for this point using $f$. With the new assignment, we recompute the weights of the clusters and we do the same thing until the main objective $L_{f}(X,Y,W)$ has converged.

The only supplement part is the function \textit{DynamicProgrammingPathFinder} that comes handy in the study of series. The goal of this function is to find the clustering assignment optimal with respect to the objective function such that it puts constraints on the number of cluster, the number of transitions and the minimal size of any block being $Min_{\phi}$ considered as an input. The function \textit{DynamicProgrammingPathFinder} will constrain the assignment and return the optimal one using Dynamic Programming by solving a recurrence relation that we detail now. 

We define $\omega_{t}(n,c)$:  the optimal total cost (cumulatively summed over the steps $\{0,...,t\}$) of assigning the point $t$ to the cluster $c$ and having used $n$ transitions over the steps $\{0,...,t\}$ among all the possible assignments of the steps $\{0,...,t\}$ that use $n$ transitions with blocks bigger than $Min_{\phi}$.

The following recurrence relation must be satisfied by $\omega_{t}(n,c)$ (proof given in Appendix \ref{app:proof2}):

\begin{subequations}\label{eq:RR2}
\begin{empheq}[left=\empheqlbrace,right=(\ref{eq:RR2})]{align*}
	\forall t < Min_{\phi},  \forall n, \forall c,\\ \omega_{t}(n,c) & =
 		\begin{cases}
    		\sum_{k=0}^{t} f(x_{k},w_{c}) & \mbox{if } n=0 \mbox{ \& } t=Min_{\phi}-1 \\
    		+ \infty & \mbox{else} \\
    	\end{cases}\\
    \forall t \geq  Min_{\phi}, \forall c,\\ 
    \omega_{t}(0,c) & = f(x_{t},w_{c}) + \omega_{t-1}(0,c)\\
    	\\
    	\forall t \geq Min_{\phi}, \forall n \neq 0, \forall c, \\ \omega_{t}(n,c) & = \min 
    	\begin{cases}
    		f(x_{t},w_{c}) + \omega_{t-1}(n,c) \\
    		 \sum\limits_{k=t-Min_{\phi}+1}^{t} f(x_{k},w_{c}) + \min\limits_{c^{'} \neq c} \omega_{t-Min_{\phi}}(n-1,c^{'}) \\
    	\end{cases}\\
\end{empheq}
\end{subequations}

The optimal cost we are interested in is $\min\limits_{\substack{ \forall c, \forall n}} \omega_{T-1}(n,c)$. When we want a transition, we look back at $Min_{\phi}$ points in the past and we check whether a transition $Min_{\phi}$ points before would have been suitable. In the opposite case, we rely only on $\omega_{t-1}(n,c)$. Solving this recurrence relation yields the minimal cost. This is done using a classical bottom-up approach. As a matter of fact, the optimal assignment is given in a linear time once this recurrence relation is solved. 

The final prediction has \textit{at most} $N$ transitions and  
\textit{at most $C$} different clusters.

\begin{algorithm}
\DontPrintSemicolon
  
  \KwInput{X, $M_{char}$, f, $C$, $N$, $Min_{\phi}$, $\epsilon$}
  \KwOutput{Y}
  \KwInit{}
  oldLoss = $\infty$ \\
  currentLoss = 0 \\
  Y = randomAssignmentTransition(C,len(X),N,$Min_{\phi}$) \\
  T = len(X) \\
  \BlankLine
  \While{abs(currentLoss-oldLoss) $>=$ $\epsilon$} 
   {
   \tcc{Until convergence}
   		\BlankLine
   		$\{w_{c}\}_{1 \leq c \leq C}$ = W $\gets$  $M_{char}$(X,Y,C,$Min_{\phi}$) \tcp*{Compute the weights}
   		\BlankLine
   		
   		Y = DynamicProgrammingPathFinder(W,X,C,N,f) \tcp*{New assignment with recurrence relation \ref{eq:RR2} }
    		
    	\BlankLine
    	oldLoss = currentLoss 
    	\BlankLine
    	currentLoss = $L_{f}$(X,Y,W)
   }
   
\BlankLine

\caption{Clustering Algorithm - Final}
\label{algo3}
\end{algorithm}

\section{Practical Considerations}
\subsection{Adjusted Rand Index}
\label{sec:ARI}
The algorithm begins with a random initialization. In order to have a stable answer, we will perform the previous algorithm 
$N_{init}$ times. We thus have $N_{init}$ different assignments. Finally, to have one final assignment, we will compute the pairwise adjusted rand index (ARI \cite{10.1007/978-3-642-04277-5_18}) between those $N_{init}$ assignments and take the one that has the highest mean ARI. We briefly review below the definition of ARI.

Let us start by considering the \textit{unadjusted} Rand Index (RI). Given two assignments $A$ and $B$, the \textit{unadjusted} RI is given by $\frac{a+b}{C_{2}^{n_{samples}}} $, where $a$ is the number of pairs of elements that are in the same set in $A$ and in the same set in $B$, $b$ the number of pairs of elements that are in different sets in $A$ and in different sets in $B$, and $C_{2}^{n_{samples}}$ the total number of possible pairs in the dataset (without ordering).

The ARI adjusts RI by removing the expected RI of random labelings. We give the exact formulation below. We note $A_{i}$ the set of index of points that have been labeled $i$ and in the same manner $B_{i}$. We note $ n_{ij}=|A_i \cap B_j| $, $a_{i} = \sum_{j} n_{i,j}$ and  $b_{j} = \sum_{i} n_{i,j}$. The ARI between the assignments $A$ and $B$ is defined as:

$$ ARI(A,B) = \frac{\sum_{ij} \binom{n_{ij}}{2} - [\sum_i \binom{a_i}{2} \sum_j \binom{b_j}{2}] / \binom{n}{2} }{ \frac{1}{2} [\sum_i \binom{a_i}{2} + \sum_j \binom{b_j}{2}] - [\sum_i \binom{a_i}{2} \sum_j \binom{b_j}{2}] / \binom{n}{2} } $$

As a result, the ARI is a score between -1.0 and 1.0 where 1.0 is the perfect match score. Random uniform label assignment have an ARI close to 0.0. 	

\subsection{Grid-Search}
\label{sec:grid_search}

Although for a given set of $N$ and $C$, the proposed algorithm \ref{algo3} together with sufficient number of the random initialization provides a scheme to effectively cluster the multi-variate series, one common piece of puzzle for unsupervised clustering is often the unknown values of the ``\textit{correct}'' $N$ and $C$ in practical applications. Typically, when higher values of $C$ and $N$ are used, lower total costs of the final clustering assignment is expected. The unbounded values of $N$ and $C$ could easily result in a structure more fragmented than the one that is desired. Hence, we need to perform a grid-search over $N$ and $C$ and yield a final answer by applying an additional criterion.

From a general perspective, the goal is to have a mean to determine a clustering pattern given by a set of $N$ and $C$ that neither leads to over fragmented clusters nor has an unacceptably high total cost. \textit{Empirically, we have taken the following procedure to give the final assignment.} We will take the most common assignment, using ARI as precised in Sec. \ref{sec:ARI}, among the plausible cluster patterns falling within a range of grid points satisfying the following criterion: \textit{A region consists of high diversity of grid points parameterized in $N$ and $C$ with convergent total cost.} 

\begin{figure}
\centering
\includegraphics[trim=2cm 1cm 2cm 2cm,clip,scale=0.35]{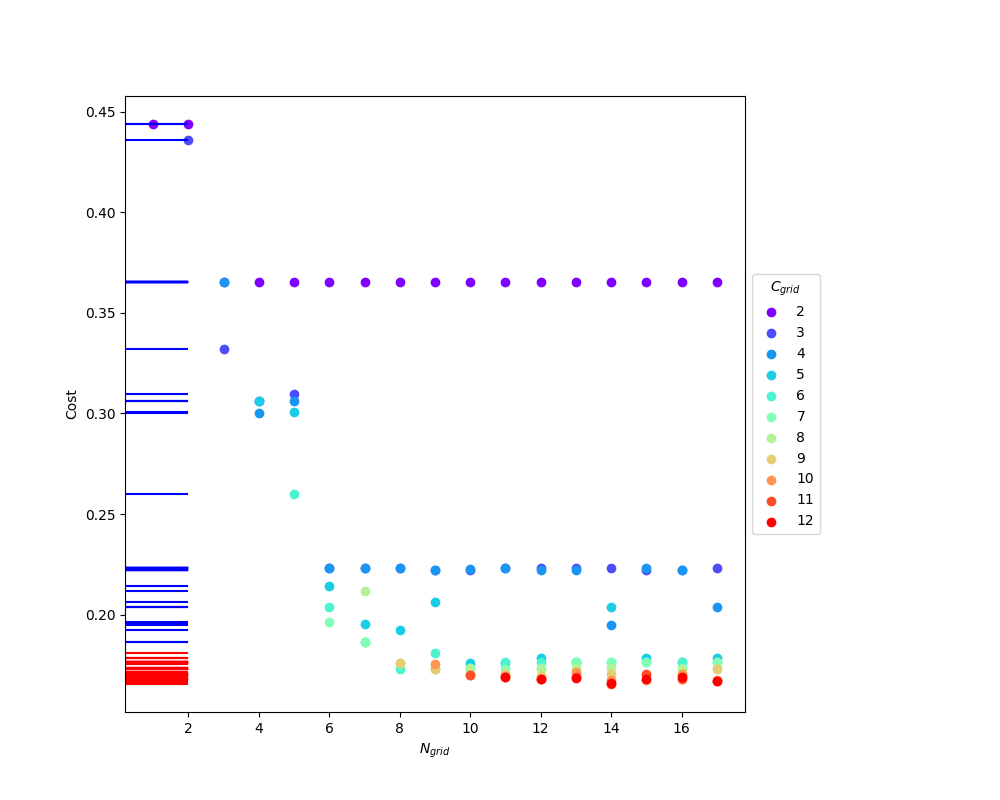}
\centering
  \caption{Grid-search example: The y-axis is the MSE cost, the x-axis is the number of transitions and the color is the number of clusters. On the left, we have the rug plot described in \ref{sec:grid_search}. The red ticks correspond to the most common region found when applied the grid-search criterion described in \ref{sec:grid_search}}
  \label{fig:ex_grid_search}
\end{figure}

The selecting criterion for the grid points giving plausible cluster patterns can be clearly depicted by plots shown in figure~\ref{fig:ex_grid_search}, where the red ticks along the y-axis indicate the region of the convergent total cost. In figure~\ref{fig:ex_grid_search}, $N_{grid}$ and $C_{grid}$ represents the ($N$,$C$) input to the algorithm (and not the number of transition/cluster of the associated assignment). On the left of the figure, we have one blue tick on the cost axis per point (rug plot) - it enables us to see the distribution of total cost from grid points. Using plots similar to figure~\ref{fig:ex_grid_search}, we can recast our selecting criterion as follows: \textit{the most dense region of the distribution of points on the cost axis that contains points of high diversity in $N$ and $C$.}

The idea behind the criterion is two-fold:
\begin{itemize}
\item \textit{the most dense region of the distribution of points on the cost axis}: we want a stable cost dense region. Indeed, when doing the grid-search many ($N_{grid}$,$C_{grid}$) pairs comes back to the same assignment (or approximately the same) yielding more or less the same cost. This is due to the fact that when starting with ($N_{grid}$,$C_{grid}$), we might have an answer with $N' \leq N_{grid}$ and $C' \leq C_{grid}$.

\item \textit{that contains points of high diversity in $N_{grid}$ and} $C_{grid}$: we want a region not biased toward one $C$ or one $N$. In the example (figure \ref{fig:ex_grid_search}), for $C=2$ the cost is stable but it is mainly due that we need more clusters to observe a change in the cost. 
\end{itemize} 

\textit{This criterion might be subjective when deciding for "the most dense" region but we did not find another criterion that gives better results for the studied datasets.} Moreover the criterion depends on the size of the grid. Indeed, when the grid gets bigger, that is going for higher $N$ and $C$, the total cost can get lower with the issue of too fragmented clustering. In such scenario, the criterion described above becomes unreliable as the selected cost range might lead to more fragmented assignments. For many applications, this is not a problem as the $N$ and $C$ can be constrained within a range based on prior information. So the algorithm will be computed on a realistic grid with respect to the studied dataset. On the other hand, it is also possible that one cannot find and define a convergent region for the total cost. In such case, one can only take the lowest cost answer as the final assignment.

\section{Application to petrophysics}
We give here an application of the algorithm for clustering petrophysical well log data. The experiments and results  given in this paper are for the specific application that we detail in this section.

In the context of sub-surface characterization, petrophysicists have recorded many measurements to infer the formation properties such as the porosity, the water saturation, the conductivity of the water-filled rock, grain density, gamma ray,... 
We name these properties \textit{logs} for the rest of the work. Moreover, petrophysicists have studied parametric equations combining these different logs in order to characterize data-points belonging to the same formation type. 
In other words, data-points belonging to the same formation type should follow these physics equations with the \textit{same} parameters.

In the case of clustering petrophysical well log data, we better understand the usefulness of the constraints considered in the algorithm:
\begin{itemize}
    \item Unsupervised: there is not known label in any petrophysical dataset
    \item Number of cluster: there is a small finite number of different mineralogy formations that makes sense for the exploitation
    \item Number of transition: the geological structure cannot be too fragmented
    \item Minimal size of a block: each block represent a formation layer that must be bigger than a prior minimal size.
\end{itemize}

One of the key physics relation that could help describing a cluster is the conductivity response of the water-filled rock ($\sigma_{o}$) constrained by the water conductivity ($\sigma_{w}$), the porosity ($\phi$), the volume fraction of clay ($f_{clay}$) and the water saturation fraction ($S_{w}$). In the literature, many equations try to study this relation. Among those, the Waxman-Smits (WS) equation (\cite{Waxman1968}) cover a wide variety of the formation types and is used in many cases in the petrophysics domain. As such, in this study, we will focus on this equation. A primary input to the Waxman-Smits (WS) equation is the volume concentration of clay exchange cations ($Q_{v}$) which is computed from the cation exchange capacity (CEC):

\begin{equation}
\label{eqn:qv}
\boxed{Q_{v} = \frac{CEC \times f_{clay} \times (1-\phi)}{\phi}}
\end{equation}

The WS equation is:

\begin{align}
\label{eqn:ws}
    \boxed{\sigma_{o} = \phi^{m} \times S_{w}^{n} \times \left( \sigma_{w} + \frac{B \times Q_{v}}{S_{w}} \right) }
    \\[-\baselineskip]
    \eqname{WS equation}
\end{align}

where $m$ and $n$ are some parameters and $B$ is a function of the temperature $T$ which is considered constant throughout the studied logs.

\begin{equation}
\label{eqn:B}
\boxed{B = \left( 1 - 0.83 \times e^{ -\sigma_{w}  \times  \left( -2.47 + 0.229  \times  ln(T)^{2} + \frac{1311}{T^{2}} \right) ^{-1} } \right)  \times  \left(-9.2431 + 2.6146  \times  \sqrt{T} \right) }
\end{equation}

We will leverage the WS equation (\ref{eqn:ws}) as a mean to characterize a cluster. Among all the variables involved in the precedent equations, the logs of $\phi$, $S_{w}$, $\sigma_{o}$, $f_{clay}$ are our input. In order to characterize a cluster, we perform a regression of $m$, $n$, $\rho_{w}$ (=$\frac{1}{\sigma_{w}}$), and CEC by minimizing the L2 loss function between the experimental (given well log) $\rho_{o}$ (=$\frac{1}{\sigma_{o}}$) and the predicted one computed with the WS equation (\ref{eqn:ws}) on the points of that cluster.

We describe here the exact values of the parameters for the algorithm \ref{algo3} in the case of the WS regression. 

\begin{itemize}
\item $X$ are the 4 data logs $X$ = [$f_{clay}$, $\phi$, $S_{w}$, $\rho_{o}$] ($d=4$). Please keep in mind this specific order of the columns of $X$.
\item $T$ is the number of data-points of the logs, $T=len(X)$.
\item $C$ is the number of different types of formation layer.
\item $N$ is still the maximum number of transitions between formation layers.
\item $Min{\phi}$ is the minimum size of a formation layer and is also the number of points that we judge enough for the WS regression (for the computation of the weights $w_{c}$ to be meaningful.
\item $N_{init}$: experimentally set to 50.
\item $M_{char}$ and $f$:
$M_{char}(X,Y,C,Min{\phi})$ for the cluster characterization and the function $f$ for the cluster affiliation are detailed in algorithm \ref{ws_algo}. The $WS_{regression}$ function is a regression using the L2-Loss over the WS equation \ref{eqn:ws}, regressing $w = [m,n,\rho_{w},CEC] $. Here, we use the same L2-Loss for the regression and for the function $f$.
The regression is performed using Scipy function optimize.fmin (that implements the downhill simplex algorithm). Finally, $WS_{forward}(x,w)$ is just a function applying the WS equation \ref{eqn:ws} with the corresponding $ m,n,\sigma_{w},CEC $  to get the predicted $\rho_{o}$.

\end{itemize}

\begin{algorithm}
\DontPrintSemicolon
$f(x,w) = || x[0] - WS_{forward}(x,w) ||^{2}$ \tcp*{L2 Loss on $\rho_{o}$}

\BlankLine

\Fn{$M_{ws,char}$(X,Y,$C$,$Min{\phi}$)}{
	W = empty array of size $C$
	\BlankLine
	\For{$c\gets1$ \KwTo $C$} { 
		tmp = $\{ X[t] \mid Y[t]=c\}$
		
		\If {len(tmp) $\geq$ $Min_{\phi}$} {
			W[c] = $WS_{regression}$(tmp)
		} 
		\Else {
		Remove cluster c
		}
	}
	
	return W
}

\caption{Cluster characterization and affiliation for WS}
\label{ws_algo}
\end{algorithm}

\section{Experiments}
\label{sec:exp_sim}
As mentioned previously, there is no dataset with groung-truth label in the adressed problem. As such, we have simulated our own well logs with a noise level close to the one present in the field well data.

In order to simulate data logs, we have first defined a way to simulate logs for a specific formation layer. For this purpose, we have fixed the expected value for $m$, $n$, $CEC$, $\sigma_{w}$ and the interval of $\phi$, $S_{w}$, $f_{clay}$ for each formation type / cluster. With these inputs, we can directly generate logs for $\phi$, $S_{w}$, and $f_{clay}$ but they will not show a realistic "time-series" correlation. Therefore, we made an adjustment to the $\phi$ and $S_{w}$ logs: we do not want points to jump between largely different values at each depth, i.e. we would like to have a smooth behavior. Moreover for $S_{w}$ the adjustment is made such as to have the water at the bottom of the formation as it is usually the case in the nature: for the same block, $S_{w}$. The fraction of clay might not need such an adjustment. As such, we have a reference signal for $\phi$, $S_{w}$ and $f_{clay}$.

Then, we added noise of different level to these logs. Indeed in the real cases, we expect the field measurements to have inherent uncertainty for $\phi$, $S_{w}$, and $f_{clay}$. Furthermore, we added noise to $m$, $n$, $CEC$ as there is a natural variation of these value even within the same formation type. More precisely, we added approximately between 5\% to 10\% error to $\phi$, $S_{w}$, $f_{clay}$ and $CEC$, a Gaussian noise with a standard deviation 0.05 to $m$ and 0.03 to $n$.

Finally, we have generated $\sigma_{o}$ log using the WS (\ref{eqn:ws}) equation with these noise-added signals as input. This gives us a way to generate realistic data logs given parameters describing a specific formation type or cluster. As such, we have generated multiple clusters with several blocks per cluster. To have a full well log, we put together the blocks either in a sharp manner (simple concatenation) or in a smooth manner (by interpolating between the blocks). The datasets named with a suffix "-smooth" present smooth transitions while the others present sharp transitions.

The grid of the grid-search (as mentioned in \ref{sec:grid_search}) is defined as $1 \leq C \leq k \times C_{true}$  and $0 \leq N \leq k \times N_{true}$  where $C_{true}$ and $N_{true}$ are the true values of respectively $C$ and $N$, and $k$ is a constant equal to approximately 1.5. The idea behind this constant $k$ is that geological priors refrain the prediction to have high $N$ and $C$ with respect to the real datasets: the grid must be "realistic".

\section{Results}

We have tested our algorithm on seven simulated datasets. We present thoroughly here the results on two of them (WS-3 and WS-3-smooth) and leave the others in the Appendix.
In table \ref{tab:results_WS}, we give numeric results over all the datasets, showing that the method present very reliable predictions looking at the ARI.

For each dataset, we first give the table of parameters that were used to generate the data along with the cluster label.

Then, we give  a scatter plot of the cost against $N$ for each cluster to see where we apply the grid-search criterion mentioned in section \ref{sec:grid_search}. The blue ticks on the scatter plot represent the rug plot. The red ticks designate the region of the cost taken to compute the final assignment by taking the most common pattern among the points of this region: this is the region described in section \ref{sec:grid_search}. $N_{grid}$ and $C_{grid}$ represents the ($N$,$C$) input to the algorithm.

Finally, we give the plots of the predictions: on each plot, the prediction on the left is either the most common answer or the lowest cost one and the prediction on the right is the ground truth. Each color represents a different cluster within the prediction (cluster 0 of the prediction and of the ground truth might have nothing in common). The y axis represents the depth and the x axis is simply for illustration purpose. We also give their confusion matrix with the ground truth.

To better appreciate the results, one must look at the prediction plots for the transitions points and at the confusion matrix for the cluster label attribution.

For the two datasets studied in this section, the parameters used to create the clusters are the same. The difference is that in WS-3-smooth the transition are smooth as described in section \ref{sec:exp_sim}.

\begin{table}
\begin{center}
 \begin{tabular}{||c c c c c||} 
 \hline
 label & m & n & $\rho_{w}$ & CEC \\ [0.5ex] 
 \hline\hline
 0 & 1.85 & 1.7 & 0.03 & 0 \\ 
 \hline
 1 & 2.0 & 2.0 & 0.03 & 0 \\
 \hline
 2 & 2.05 & 2.0 & 0.029 & 30 \\
 \hline
 3 & 2.3 & 2.1 & 0.031 & 0 \\
 \hline
 4 & 2.5 & 2.2 & 0.049 & 80 \\
 \hline
 5 & 2.0 & 2.5 & 0.05 & 0 \\
 \hline
 6 & 2.0 & 1.9 & 0.05 & 0 \\
 \hline
 7 & 2.1 & 2.1 & 0.051 & 45 \\ 
 \hline
\end{tabular}
\end{center}
\caption{Parameters for WS-0}
\end{table}

\begin{table}
\begin{center}
 \begin{tabular}{||c c c c c||} 
 \hline
  Datasets & Prediction & $cost_{pred}$ & $cost_{true}$ & ARI  \\ [0.5ex] 
 \hline\hline
 
 WS-1 & \textbf{Most common} & \textbf{0.170} & 0.170 & \textbf{0.76} \\ 
	 & Lowest cost & 0.169 & 0.170 & 0.72 \\ 
 
 \hline
 WS-2 & \textbf{Most common} & \textbf{0.161} & 0.165 & \textbf{0.91} \\ 
  	& Lowest cost & 0.159 & 0.165 & 0.79 \\ 
 
 \hline
 WS-2-smooth & \textbf{Most common} & \textbf{0.164} & 0.163 & \textbf{0.82} \\ 
	 & Lowest cost & 0.161 & 0.163 & 0.68 \\   
 
 \hline
 WS-3 & \textbf{Most common} & \textbf{0.170} & 0.166 & \textbf{0.81} \\ 
	 & Lowest cost & 0.166 & 0.166 & 0.73 \\ 
	 
 \hline
 WS-3-smooth & Most common & 0.173 & 0.164 & 0.71 \\ 
	 & \textbf{Lowest cost} & \textbf{0.167} & 0.164 & \textbf{0.84} \\

 \hline
\end{tabular}
\end{center}
\caption{Results for simulated data. \textit{For each dataset, we have a line for the most common answer as defined in section \ref{sec:grid_search} and a line for the lowest cost answer. $cost_{pred}$ is the cost of the prediction, $cost_{true}$ is the cost of the ground truth answer, ARI is the Adjusted Rand Index between the prediction and the ground truth as defined in section \ref{sec:ARI}. In bold, we have the answer that have the highest ARI.}}
\label{tab:results_WS}
\end{table}

\begin{itemize}
    \item WS-3: in the most common answer, the cluster 1 and 2 are put together as well as 6 and 7. However, when we look at the parameters of these clusters, there are really close. Therefore, the most common prediction is still very good with respect to the high noise present in the simulation. The lowest cost answer is too fragmented as expected.
    \item WS-3-smooth: the only difference with WS-3 is that the cluster 3 is mixed with the cluster 5. Even though it can be explained with the difference in $\rho_{w}$ of these two clusters, we here have a case where the lowest cost answer gives a (much) better prediction than the most common one.

\end{itemize}

\section{Conclusions}

Given a way to characterize a cluster and to quantify the mismatch of a point to a cluster, we have designed an algorithm that enables the clustering, in an unsupervised manner, of any multi-variate series. This algorithm, based on dynamic programming, allows the user to incorporate constraints on the number of clusters, the number of transitions as well as the minimal size of a block in order to have an optimal assignment that minimizes the given objective while controlling the sparsity of the answer.

\bibliographystyle{unsrt}  
\bibliography{references}

\newpage

\begin{figure}[h!]
\centering
    \includegraphics[trim=2cm 1cm 4cm 2cm,clip,scale=0.30]{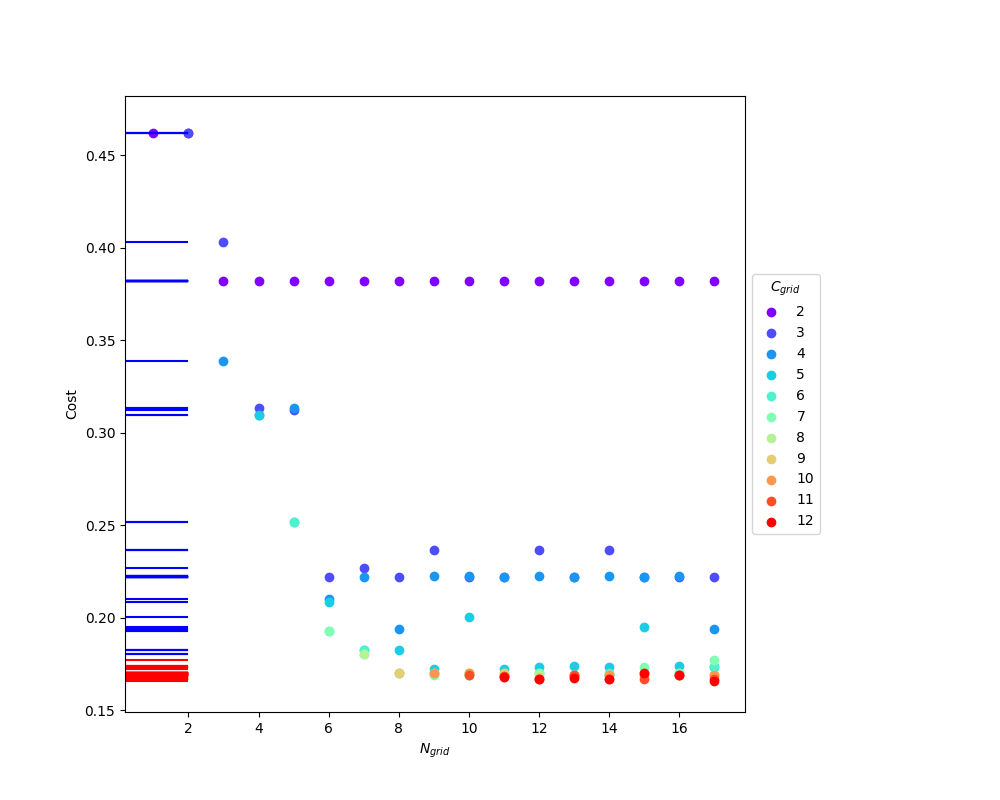}
 \caption{Grid-search criterion for WS-3\\ Most common: $N_{grid}=8$, $C_{grid}=6$}
\end{figure}

\begin{figure}[h!]
\centering
  \begin{subfigure}[b]{0.49\textwidth}
    \includegraphics[trim=4cm 2cm 2cm 2cm,clip,width=1\linewidth]{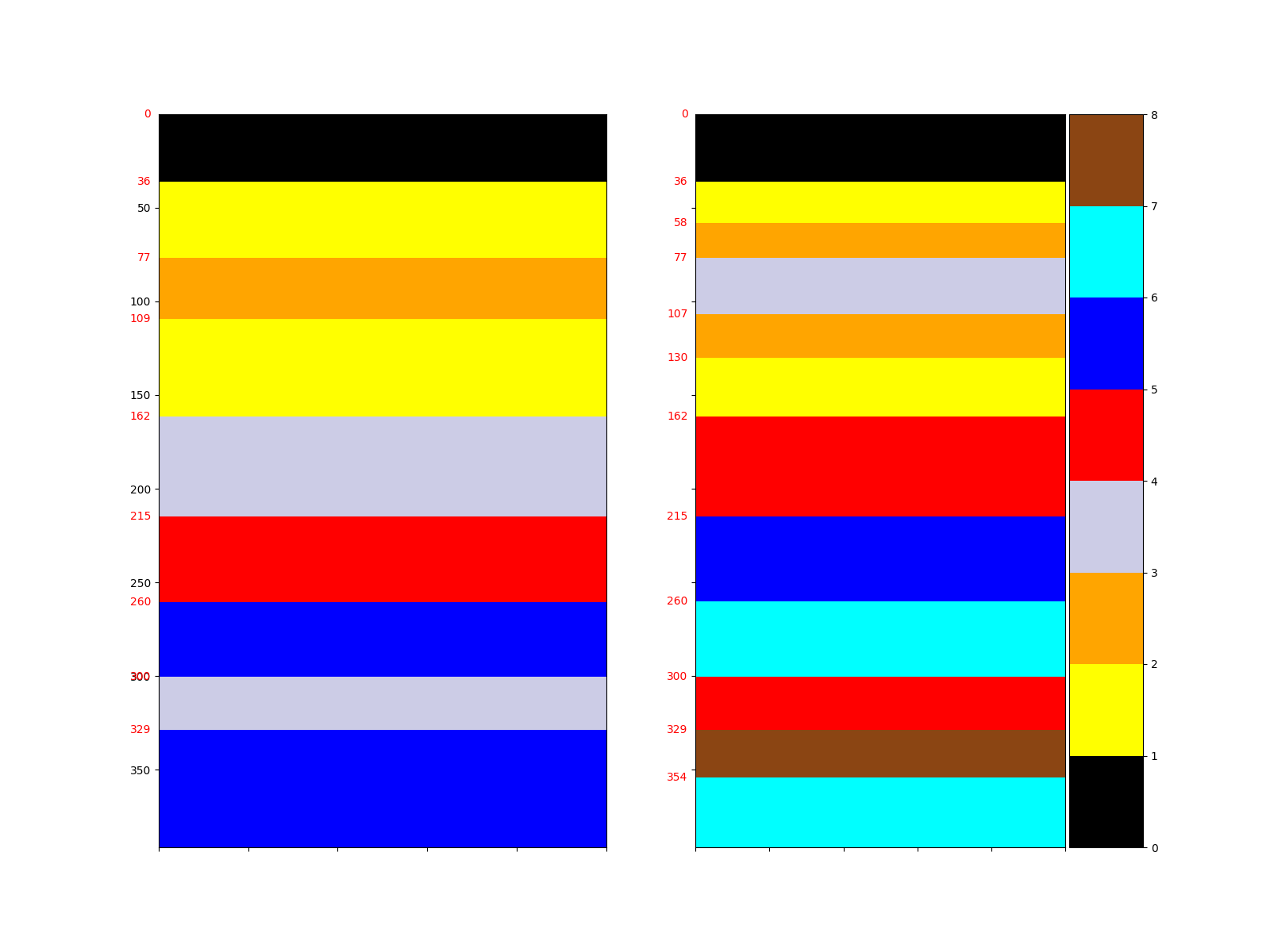}
    \caption{Most common}
  \end{subfigure}
  \begin{subfigure}[b]{0.49\textwidth}
    \includegraphics[trim=4cm 2cm 2cm 2cm,clip,width=1\linewidth]{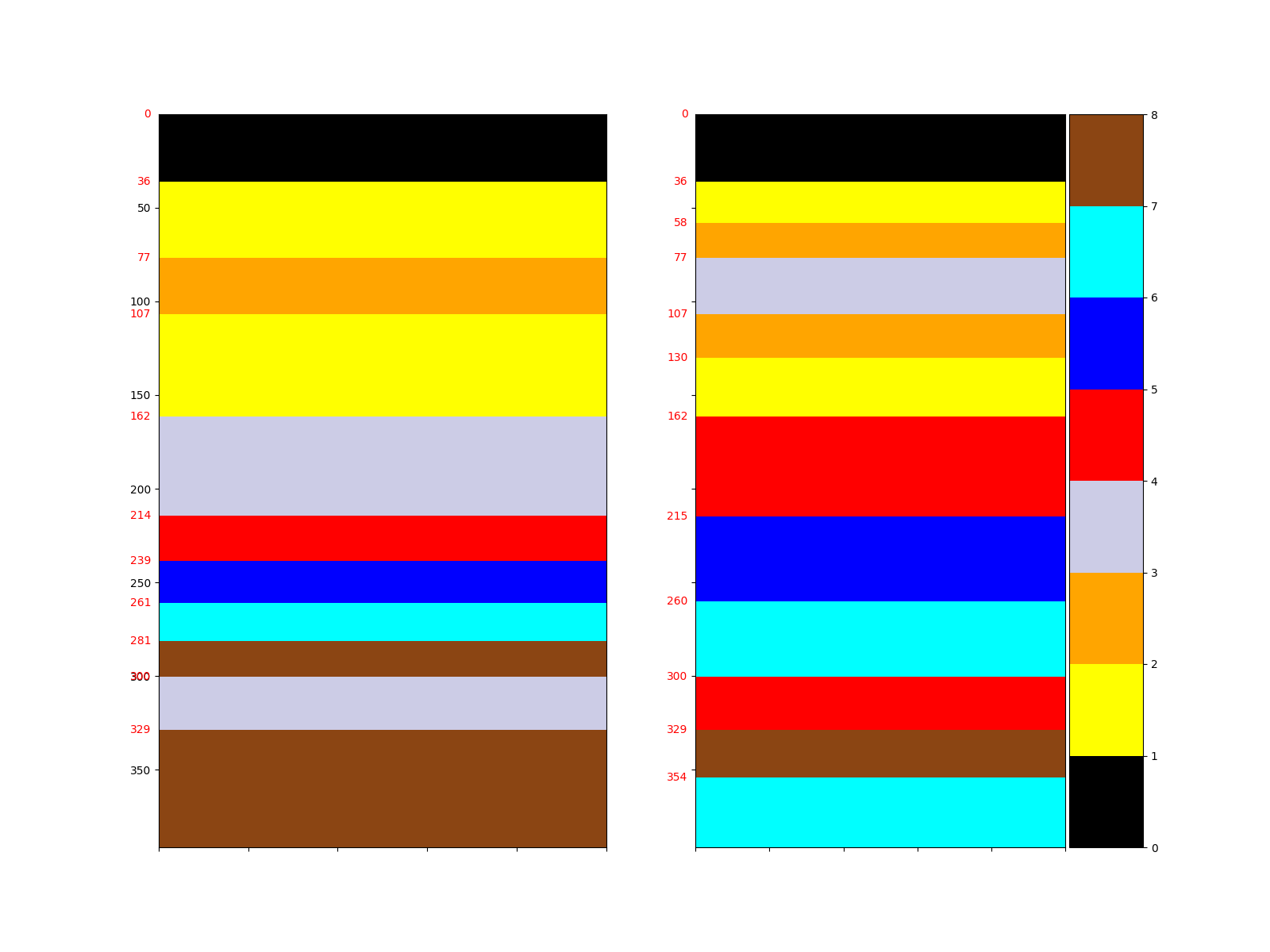}
    \caption{Lowest cost}
  \end{subfigure}
  \caption{Predictions for WS-3}
\end{figure}

\begin{figure}[h!]
\centering
  \begin{subfigure}[b]{0.49\textwidth}
    \includegraphics[trim=0cm 0cm 0cm 2cm,width=1\linewidth]{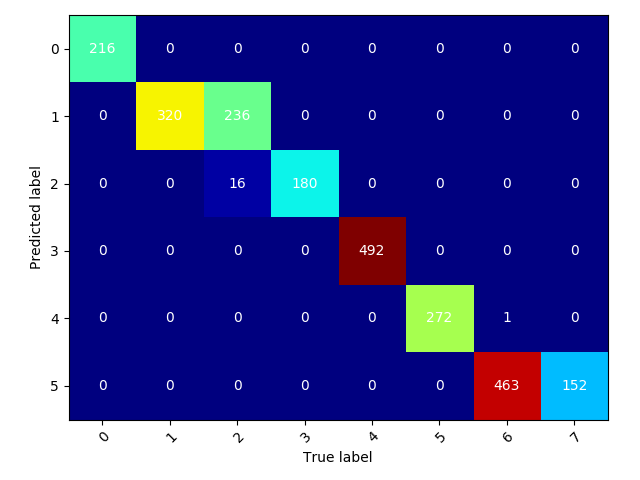}
    \caption{Most common}
  \end{subfigure}
  \begin{subfigure}[b]{0.49\textwidth}
    \includegraphics[trim=0cm 0cm 0cm 2cm,width=1\linewidth]{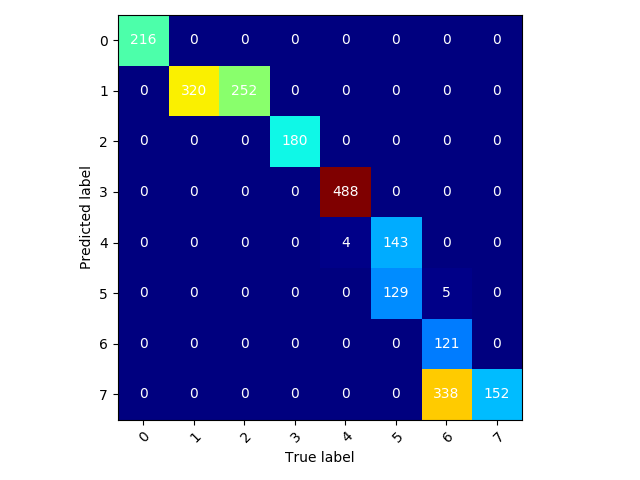}
    \caption{Lowest cost}
  \end{subfigure}
  \caption{Confusion matrices for WS-3}
\end{figure}

\begin{figure}[h!]
    \centering
    \begin{subfigure}[t]{0.6\textwidth}
        \centering
        \includegraphics[trim=2cm 1cm 4cm 2cm,clip,scale=0.30]{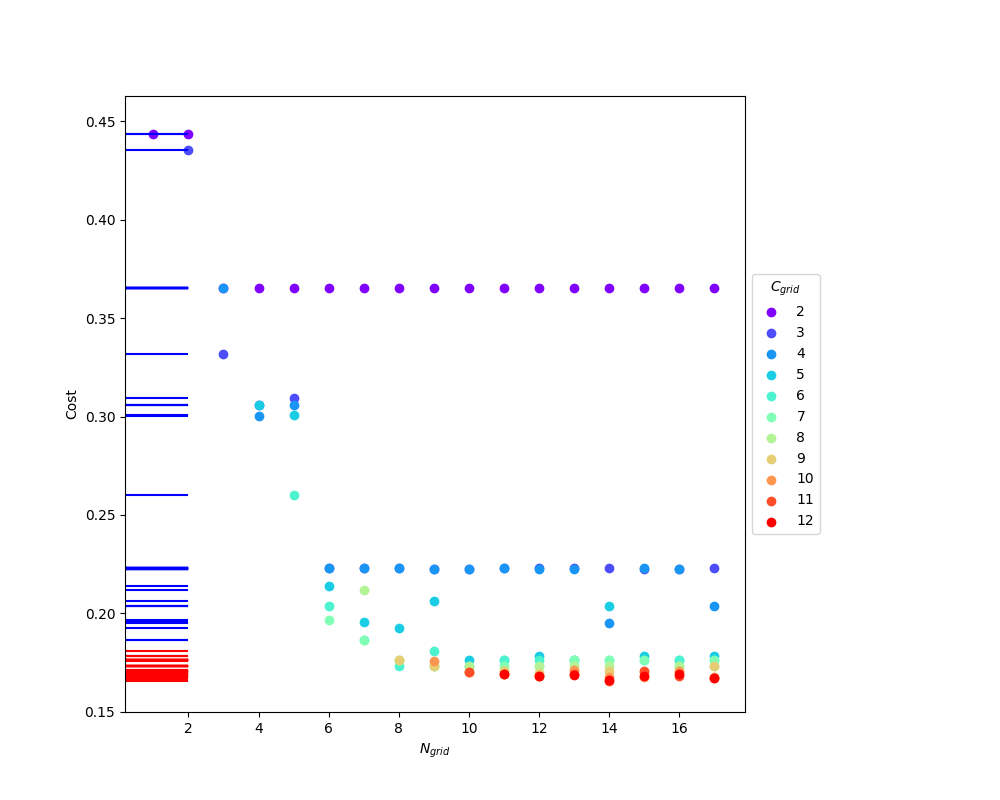}
        \caption{Grid-search criterion \\ Most common: $N_{grid}=16$, $C_{grid}=7$}
    \end{subfigure}
    \caption{Grid-search results for WS-3-smooth}
\end{figure}

\begin{figure}[h!]
\centering
  \begin{subfigure}[b]{0.49\textwidth}
    \includegraphics[trim=4cm 2cm 2cm 2cm,clip,width=1\linewidth]{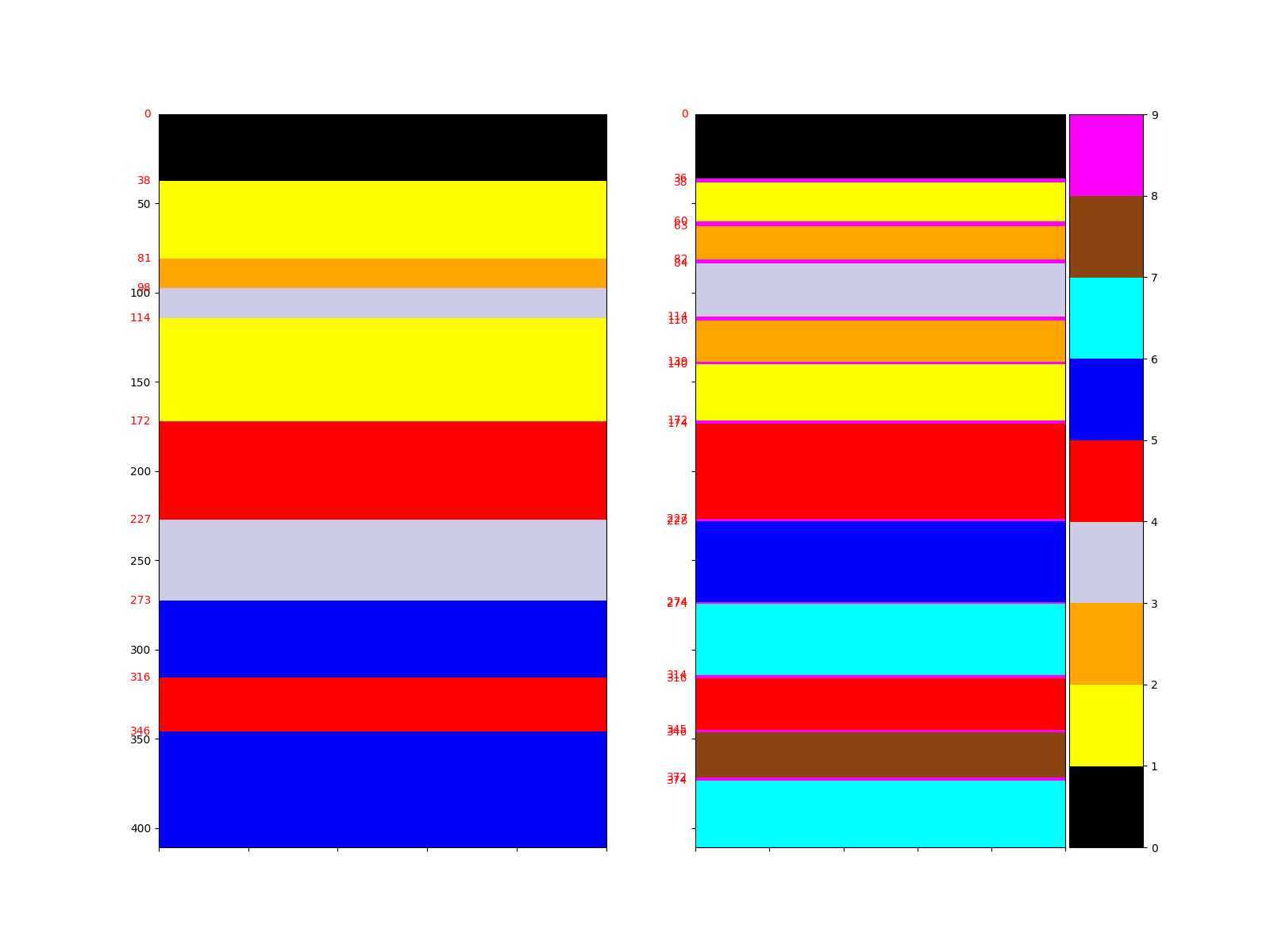}
    \caption{Most common}
  \end{subfigure}
  \begin{subfigure}[b]{0.49\textwidth}
    \includegraphics[trim=4cm 2cm 2cm 2cm,clip,width=1\linewidth]{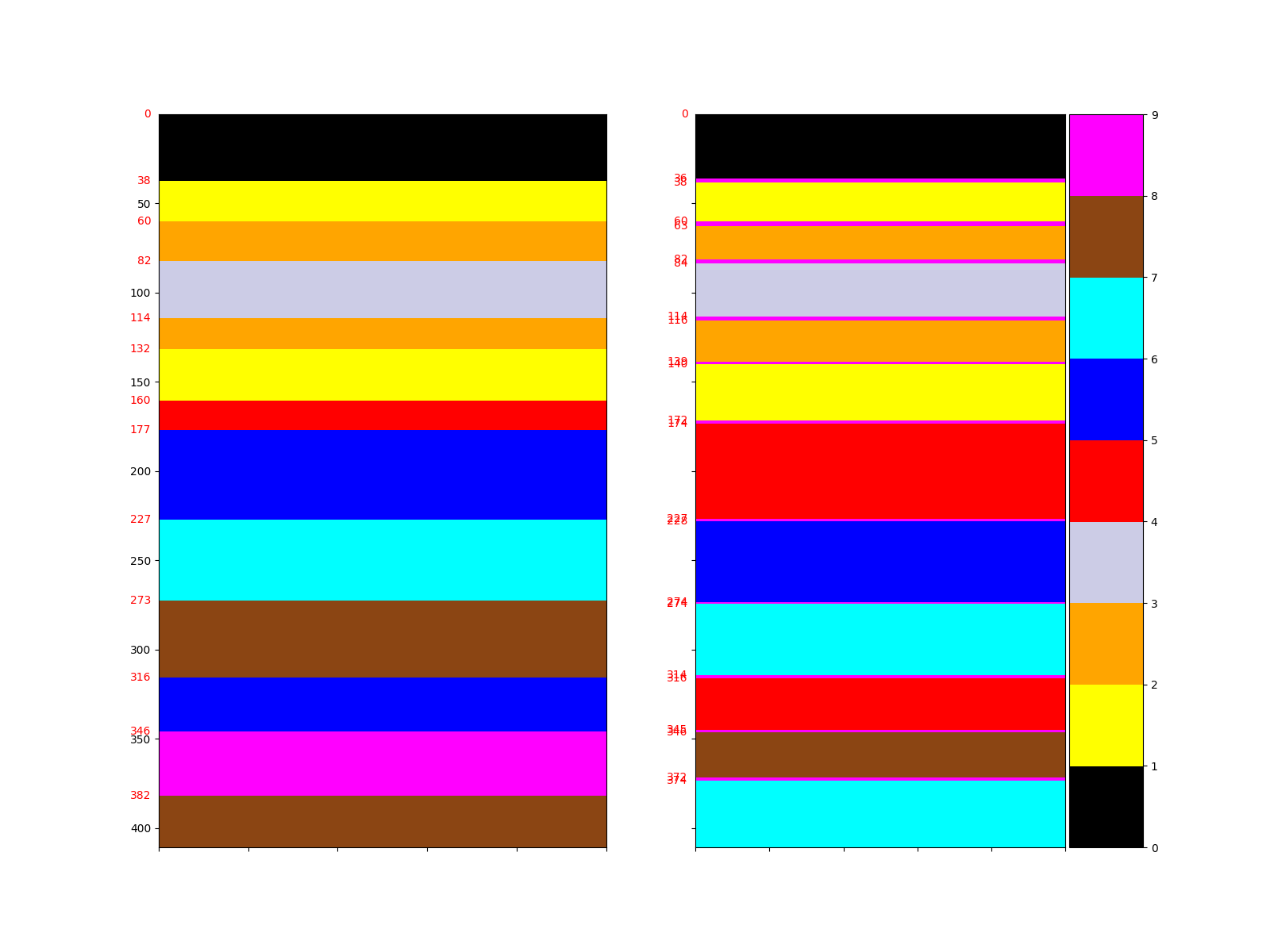}
    \caption{Lowest cost}
  \end{subfigure}
  \caption{Predictions for WS-3-smooth}
\end{figure}

\begin{figure}[h!]
\centering
  \begin{subfigure}[b]{0.49\textwidth}
    \includegraphics[trim=0cm 0cm 0cm 2cm,width=1\linewidth]{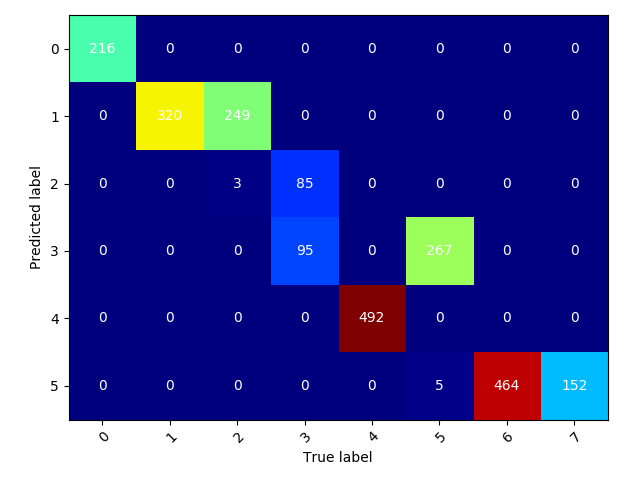}
    \caption{Most common}
  \end{subfigure}
  \begin{subfigure}[b]{0.49\textwidth}
    \includegraphics[trim=0cm 0cm 0cm 2cm,width=1\linewidth]{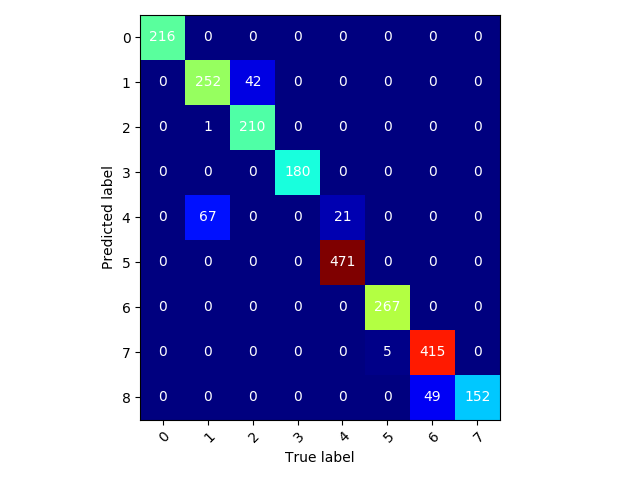}
    \caption{Lowest cost}
  \end{subfigure}
  \caption{Confusion matrices for WS-3-smooth}
\end{figure}

\begin{appendices}
\section{Proof of the Recurrence Relation}
\label{app:proof2}

We give here the proof that $\omega_{t}(n,c)$ must satisfy \ref{eq:RR2}.

We recall the definition of $\omega_{t}(n,c)$:  the optimal total cost (cumulatively summed over the timesteps $\{0,...,t\}$) of assigning the point t to the cluster c and having used n transitions over the timesteps $\{0,...,t\}$ among all the possible assignments of the timesteps $\{0,...,t\}$ that use n transitions such that all blocks should be bigger than $Min_{\phi}$:

We name $t_{k}$ ($1 \leq k \leq n$) the n transitions points and $t_0=0$ and $t_{n+1}=t+1$.
We name $Y_{i}$ the cluster label for timestep $i$.
Finally, we call $c_{k}$ ($1 \leq k \leq c$) the cluster label $k$.

\begin{equation}
\omega_{t}(n,c) = \min_{
	\substack{0=t_{0} < t_{1} < ... < t_{n} < t_{n+1} - Min_{\phi}+ 1 < t_{n+1}=t+1\\
		Y_{t_{0}}=Y_{t_{0}+1}=...\neq Y_{t_{1}} = Y_{t_{1}+1} = ... \neq Y_{t_{n-1}} = ...  = c^{'} \neq Y_{t_{n}} = ....= Y_{t} = c  \\
		\forall k \in \{0,..,n\}, \; t_{k+1}-t_{k} \geq Min_{\phi}
}}
\sum_{k=0}^{n} \sum_{s=t_{k}}^{t_{k+1}-1} f(x_{s},c_{s})
\end{equation}

We omit "$\omega_{t}(n,c) =$" in the next equation lines for more readability.

If we take out the group for the group $t_{n}$ from the double sum, we have:
\begin{equation}
\omega_{t}(n,c)=\min_{
\substack{0=t_{0} < t_{1} < ... < t_{n} < t_{n+1} - Min_{\phi}+ 1 < t_{n+1}=t+1\\
Y_{t_{0}}=Y_{t_{0}+1}=...\neq Y_{t_{1}} = Y_{t_{1}+1} = ... \neq Y_{t_{n-1}} = ...  = c^{'} \neq Y_{t_{n}} = ....= Y_{t} = c \\
 \forall k \in \{0,..,n\}, \; t_{k+1}-t_{k} \geq Min_{\phi}
}}
\sum_{s=t_{n}}^{t} f(x_{s},c_{s}) +  \sum_{k=0}^{n-1} \sum_{s=t_{k}}^{t_{k+1}-1} f(x_{s},c_{s}),
\nonumber
\end{equation}
or taking out the point at the step $t$ from the first sum:
\begin{equation}
\label{eq:recurrence_derivation1}
= f(x_{t},c) +  \min_{
	\substack{0=t_{0} < t_{1} < ... < t_{n} < t_{n+1} - Min_{\phi}+ 1 < t_{n+1}=t+1\\
		Y_{t_{0}}=Y_{t_{0}+1}=...\neq Y_{t_{1}} = Y_{t_{1}+1} = ... \neq Y_{t_{n-1}} = ...  = c^{'} \neq Y_{t_{n}} = ....= Y_{t-1} = c \\
		\forall k \in \{0,..,n\}, \; t_{k+1}-t_{k} \geq Min_{\phi}
}}
\sum_{s=t_{n}}^{t-1} f(x_{s},c_{s}) +  \sum_{k=0}^{n-1} \sum_{s=t_{k}}^{t_{k+1}-1} f(x_{s},c_{s})
\end{equation}

\begin{itemize}
\item \textbf{Case 1}

If the best $n^{th}$ transition is strictly before the last $Min_{\phi}$ points, \textit{i.e.}, $t_{n} < t-Min_{\phi}+1$ and $t-t_{n} >= Min_{\phi}$, we can safely shift the $t_{n+1}$ from $t+1$ to $t$ for all the summation terms in Eq.~\eqref{eq:recurrence_derivation1} without violating minimal block size constraint. We would have in this case:

\begin{equation}
f(x_{t},c) +  \min_{
\substack{0=t_{0} < t_{1} < ... < t_{n} < t_{n+1} - Min_{\phi}+ 1 < t_{n+1}=t\\
Y_{t_{0}}=Y_{t_{0}+1}=...\neq Y_{t_{1}} = Y_{t_{1}+1} = ... \neq Y_{t_{n-1}} = ...  = c^{'} \neq Y_{t_{n}} = ....= Y_{t-1} = c \\
 \forall k \in \{0,..,n\}, \; t_{k+1}-t_{k} \geq Min_{\phi}
}}
 \sum_{s=t_{n}}^{t_{n+1}-1} f(x_{s},c_{s}) +  \sum_{k=0}^{n-1} \sum_{s=t_{k}}^{t_{k+1}-1} f(x_{s},c_{s})
 \nonumber
\end{equation}
or in other words regrouping the sums:
$$ f(x_{t},c) +  \min_{
\substack{0=t_{0} < t_{1} < ... < t_{n} < t_{n+1} - Min_{\phi}+ 1 < t_{n+1}=t\\
Y_{t_{0}}=Y_{t_{0}+1}=...\neq Y_{t_{1}} = Y_{t_{1}+1} = ... \neq Y_{t_{n-1}} = ...  = c^{'} \neq Y_{t_{n}} = ....= Y_{t-1} = c \\
 \forall k \in \{0,..,n\}, \; t_{k+1}-t_{k} \geq Min_{\phi}
}}
\sum_{k=0}^{n} \sum_{s=t_{k}}^{t_{k+1}-1} f(x_{s},c_{s})
$$

which is equal to:
\begin{equation}
\omega_{t}(n,c) = f(x_{t},c) + \omega_{t-1}(n,c)
\end{equation}

\item \textbf{Case 2}
If the best $n^{th}$ transition is exactly at the point $t_n = t-Min_{\phi}+1$, we can take out the first sum. We would have in this case:
\begin{equation}
\nonumber
\omega_{t}(n,c) = \sum_{s=t-Min_{\phi}+1}^{t} f(x_{s},c_{s}) + \min_{
\substack{0=t_{0} < t_{1} < ... < t_{n-1} < t_{n}-Min_{\phi}+1 < t_{n}=t-Min_{\phi}+1\\
Y_{t_{0}}=Y_{t_{0}+1}=...\neq Y_{t_{1}} = Y_{t_{1}+1} = ... \neq Y_{t_{n-1}} = ...  = c^{'} \neq c \\
 \forall k \in \{0,..,n-1\}, \; t_{k+1}-t_{k} \geq Min_{\phi}
}}
 \sum_{k=0}^{n-1} \sum_{s=t_{k}}^{t_{k+1}-1} f(x_{s},c_{s})
\end{equation}

which is equal to (with $c^{'}\neq c$):
\begin{equation}
\omega_{t}(n,c) =  \sum_{s=t-Min_{\phi}+1}^{t} f(x_{s},c_{s}) + \omega_{t-Min_{\phi}}(n-1,c^{'})
\end{equation}
\end{itemize}

Finally, by realizing that the true $\omega_t(n,c)$ can be obtained through the path either from Case 1 or from Case 2, we reach the following equation as the selection criterion for the recurrence relation at the step $t$:
\begin{empheq}[]{align*}
\omega_{t}(n,c) & = \min 
    	\begin{cases}
    		f(x_{t},w_{c}) + \omega_{t-1}(n,c) \\
    		\sum\limits_{s=t-Min_{\phi}+1}^{t} f(x_{s},w_{c}) + \min\limits_{c^{'} \neq c} \omega_{t-Min_{\phi}}(n-1,c^{'}) \\
    	\end{cases}\\
\end{empheq}
which is exactly the last equation in \ref{eq:RR2}.

Moreover, we need to initialize, for $t < Min_{\phi}$ ($\forall n, \forall c$) and n=0 ($\forall t, \forall c$) in order to start this recurrence:

\begin{itemize}
\item $t < Min_{\phi}$ the only case that can have a block bigger (in this case equal) to $Min_{\phi}$ is for $t=Min_{\phi}-1$ and n=0,
\item n=0, we can have only one cluster,
\item in all other cases, the cost should be $+\infty$
\end{itemize}

\section{Other Datasets}
Here we present the clustering results for all the testing examples, beside WS-3 and WS-3-smooth, in the same fashion indicated in themain text.
\newpage

\begin{figure}
\begin{floatrow}
\ffigbox{%
 \includegraphics[trim=2cm 1cm 4cm 2cm,clip,scale=0.3]{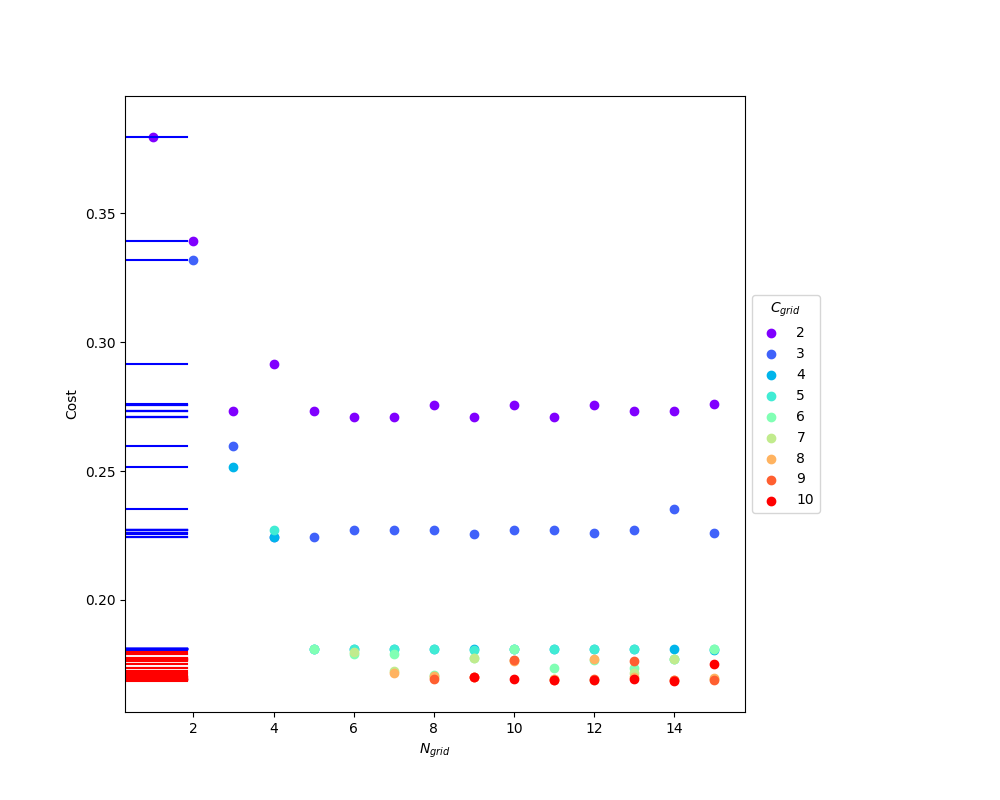}
}{%
  \caption{Grid-search criterion for WS-1 \\ Most common: $N_{grid}=14$, $C_{grid}=10$}%
}
\capbtabbox{%

 \begin{tabular}{||c c c c c||} 
 \hline
 label & m & n & $\rho_{w}$ & CEC \\ [0.5ex] 
 \hline\hline
 0 & 2.1 & 2.3 & 0.052 & 0 \\ 
 \hline
 1 & 1.9 & 1.8 & 0.05 & 0 \\
 \hline
 2 & 1.8 & 1.75 & 0.052 & 0 \\
 \hline
 3 & 2.05 & 1.9 & 0.05 & 0 \\
 \hline
 4 & 2.2 & 2.0 & 0.048 & 20 \\
 \hline
 5 & 2.4 & 2.1 & 0.048 & 60 \\
 \hline
\end{tabular}

}{%
  \caption{Parameters for WS-1}%
}
\end{floatrow}
\end{figure}

\begin{figure}
\centering
  \begin{subfigure}[b]{0.49\textwidth}
    \includegraphics[trim=4cm 2cm 2cm 2cm,clip,width=1\linewidth]{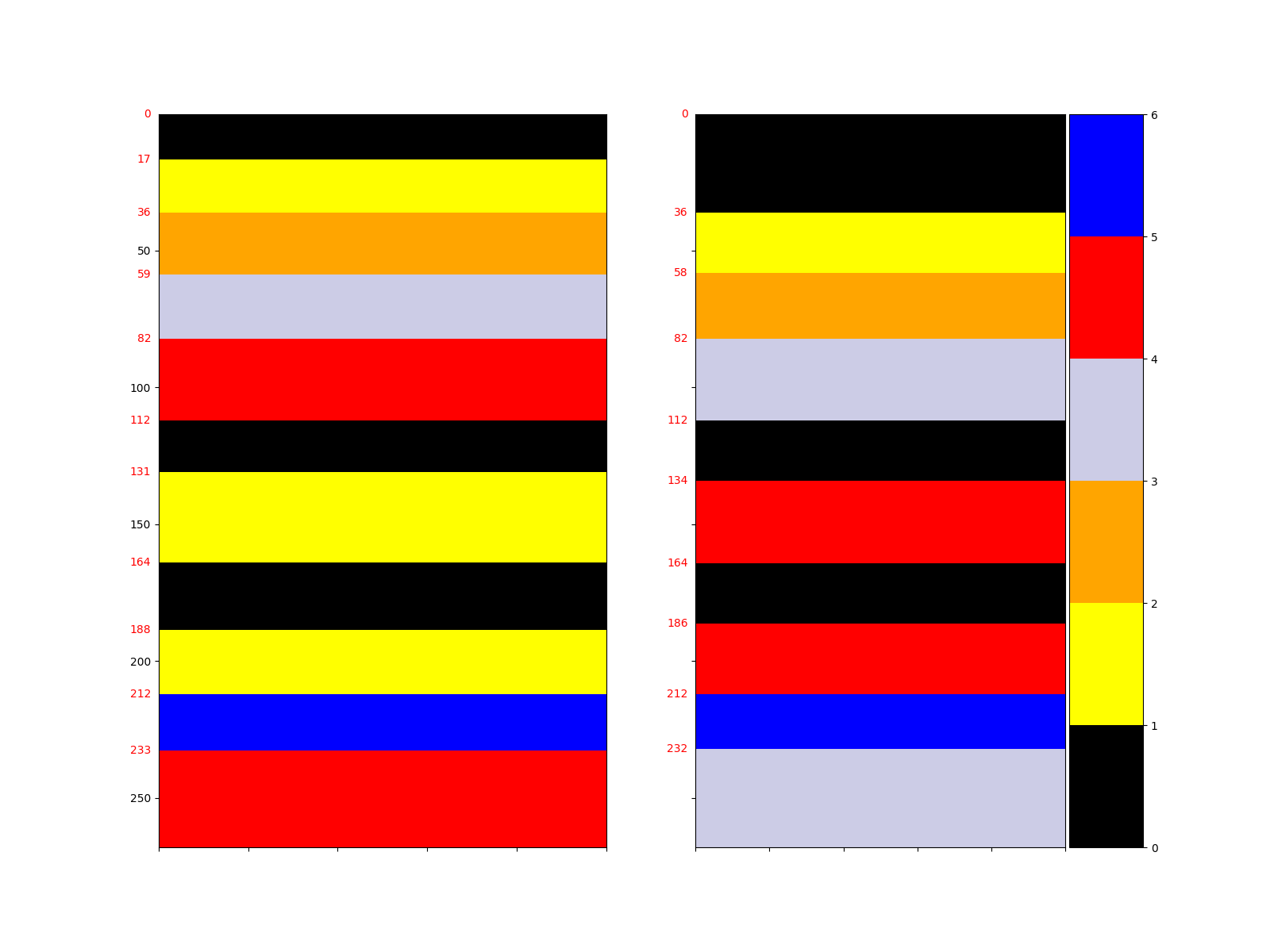}
    \caption{Most common}
  \end{subfigure}
  \begin{subfigure}[b]{0.49\textwidth}
    \includegraphics[trim=4cm 2cm 2cm 2cm,clip,width=1\linewidth]{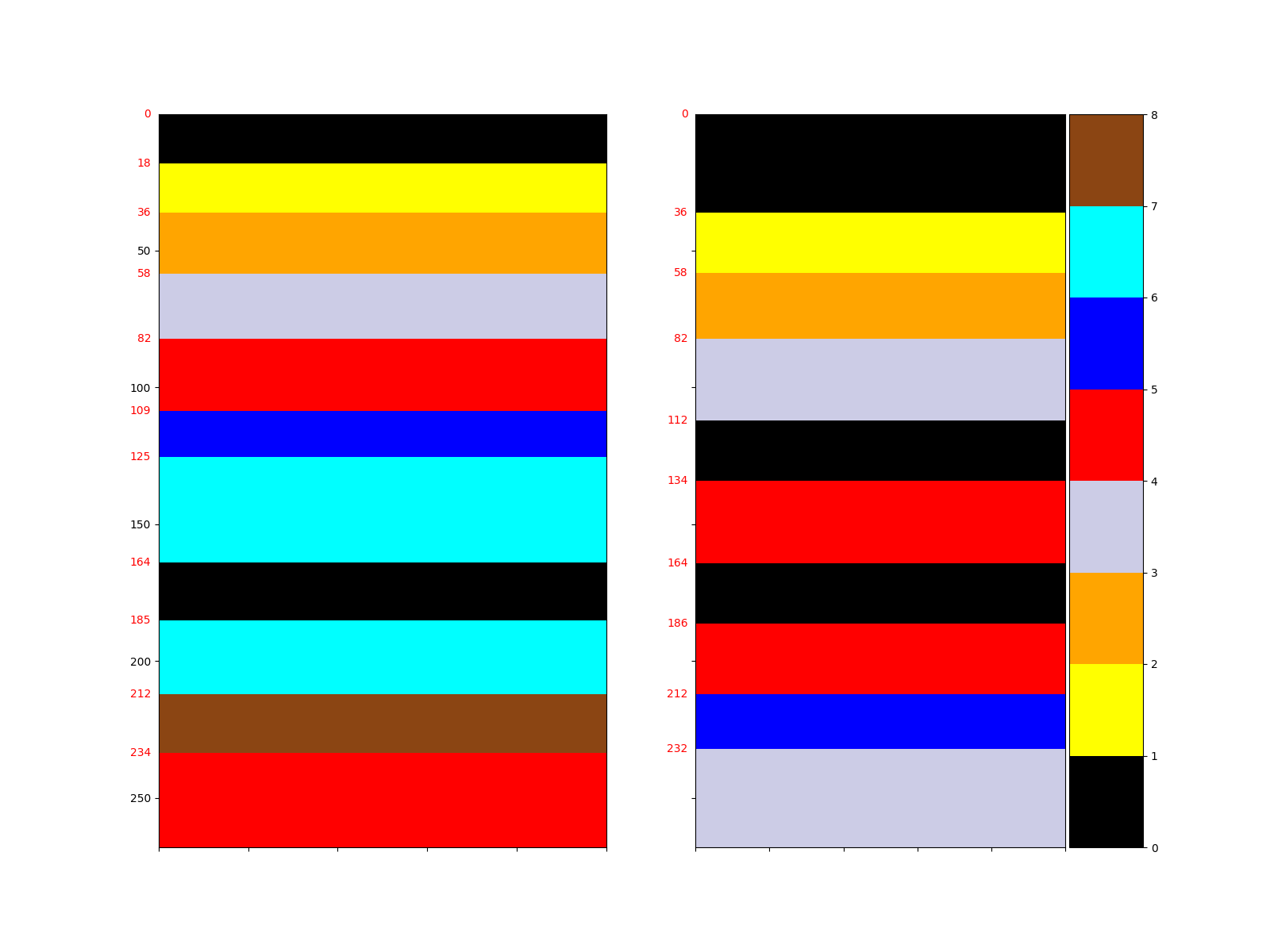}
    \caption{Lowest cost}
  \end{subfigure}
  \caption{Predictions for WS-1}
\end{figure}

\begin{figure}
\centering
  \begin{subfigure}[b]{0.49\textwidth}
    \includegraphics[trim=0cm 0cm 0cm 2cm,width=1\linewidth]{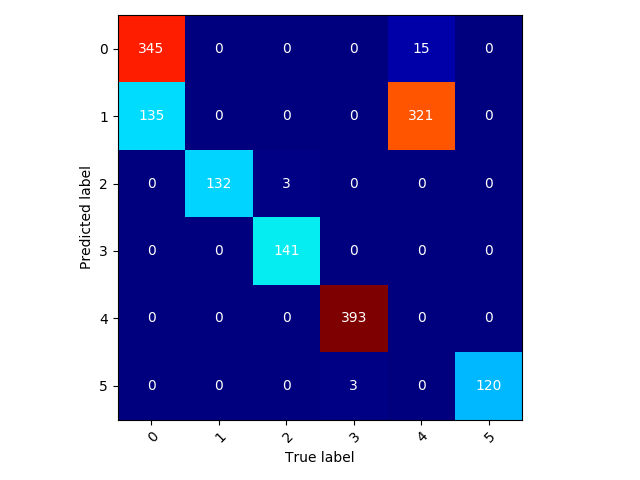}
    \caption{Most common}
  \end{subfigure}
  \begin{subfigure}[b]{0.49\textwidth}
    \includegraphics[trim=0cm 0cm 0cm 2cm,width=1\linewidth]{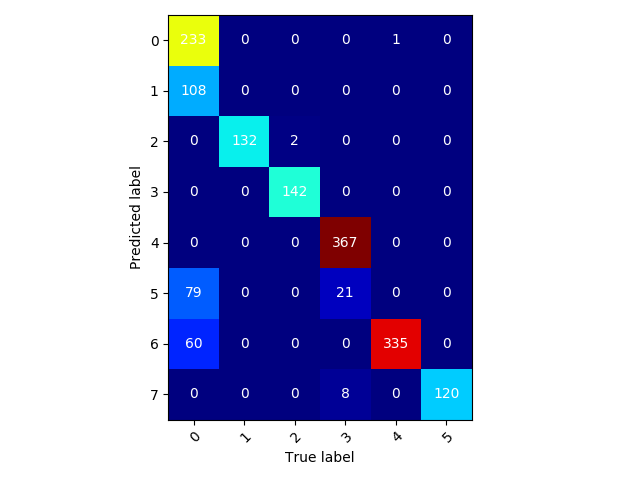}
    \caption{Lowest cost}
  \end{subfigure}
  \caption{Confusion matrices for WS-1}
\end{figure}

\begin{figure}
\begin{floatrow}
\ffigbox{%
        \includegraphics[trim=2cm 1cm 4cm 2cm,clip,scale=0.30]{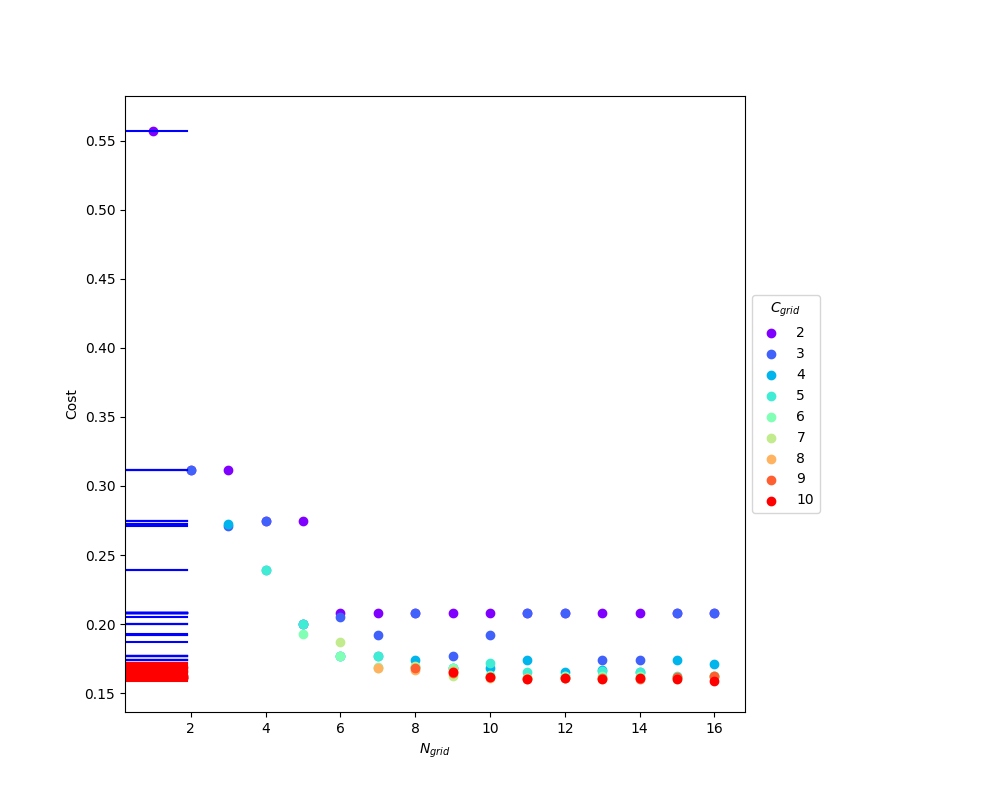}
}{%
   \caption{Grid-search criterion for WS-2 \\ Most common: $N_{grid}=14$, $C_{grid}=8$}%
}
\capbtabbox{%

 \begin{tabular}{||c c c c c||} 
 \hline
 label & m & n & $\rho_{w}$ & CEC \\ [0.5ex] 
 \hline\hline
 0 & 1.85 & 1.8 & 0.052 & 0 \\ 
 \hline
 1 & 2.1 & 2.0 & 0.052 & 0 \\
 \hline
 2 & 2.4 & 2.3 & 0.049 & 80 \\
 \hline
 3 & 1.9 & 2.0 & 0.051 & 0 \\
 \hline
 4 & 2.0 & 1.95 & 0.051 & 30 \\
 \hline
 5 & 2.0 & 2.5 & 0.05 & 0 \\ [1ex] 
 \hline
\end{tabular}

}{%
  \caption{Parameters for WS-2}%
}
\end{floatrow}
\end{figure}

\begin{figure}[h!]
\centering
  \begin{subfigure}[b]{0.49\textwidth}
    \includegraphics[trim=4cm 2cm 2cm 2cm,clip,width=1\linewidth]{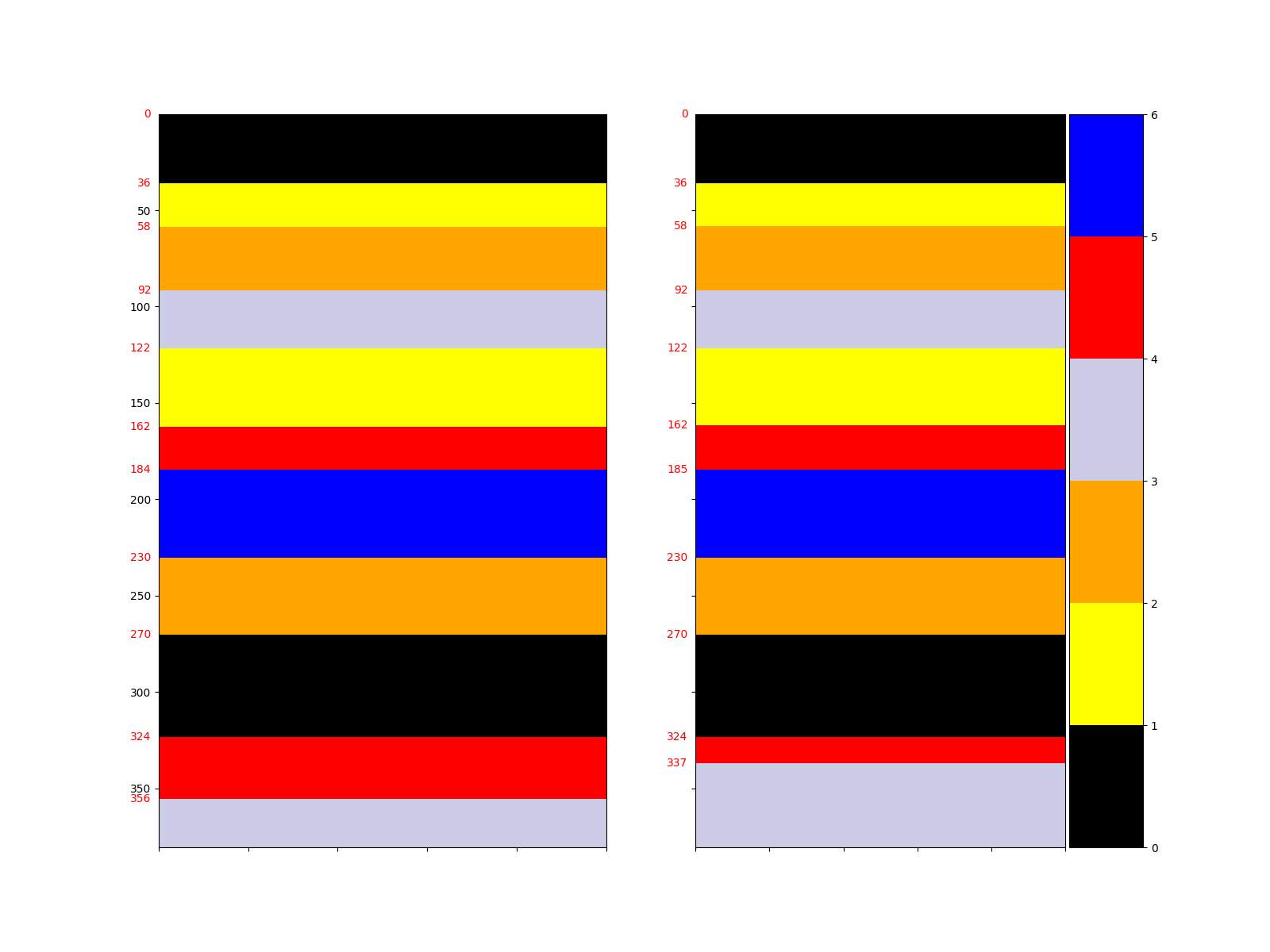}
    \caption{Most common}
  \end{subfigure}
  \begin{subfigure}[b]{0.49\textwidth}
    \includegraphics[trim=4cm 2cm 2cm 2cm,clip,width=1\linewidth]{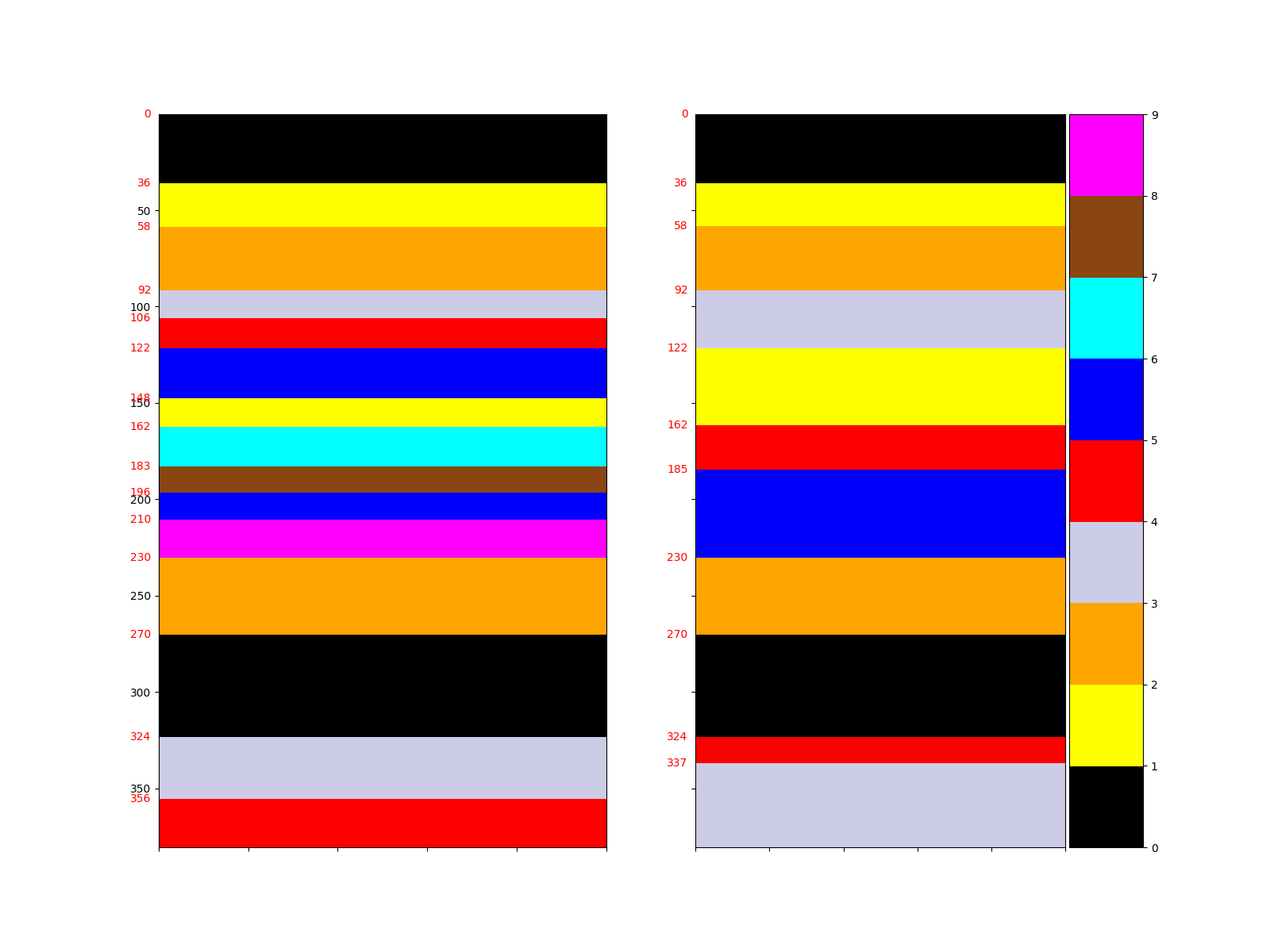}
    \caption{Lowest cost}
  \end{subfigure}
  \caption{Predictions for WS-2}
\end{figure}

\begin{figure}[h!]
\centering
  \begin{subfigure}[b]{0.49\textwidth}
    \includegraphics[trim=0cm 0cm 0cm 2cm,width=1\linewidth]{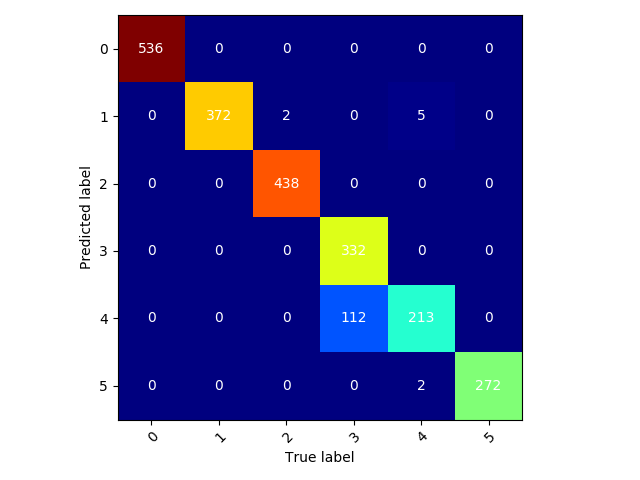}
    \caption{Most common}
  \end{subfigure}
  \begin{subfigure}[b]{0.49\textwidth}
    \includegraphics[trim=0cm 0cm 0cm 2cm,width=1\linewidth]{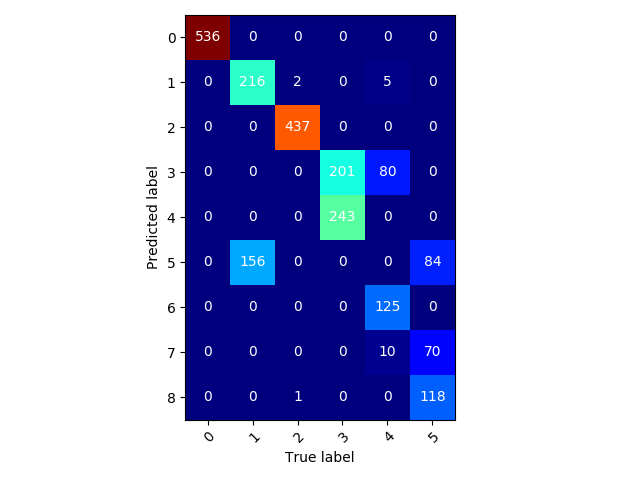}
    \caption{Lowest cost}
  \end{subfigure}
  \caption{Confusion matrices for WS-2}
\end{figure}

\begin{figure}
\begin{floatrow}
\ffigbox{%
      \includegraphics[trim=2cm 1cm 4cm 2cm,clip,scale=0.30]{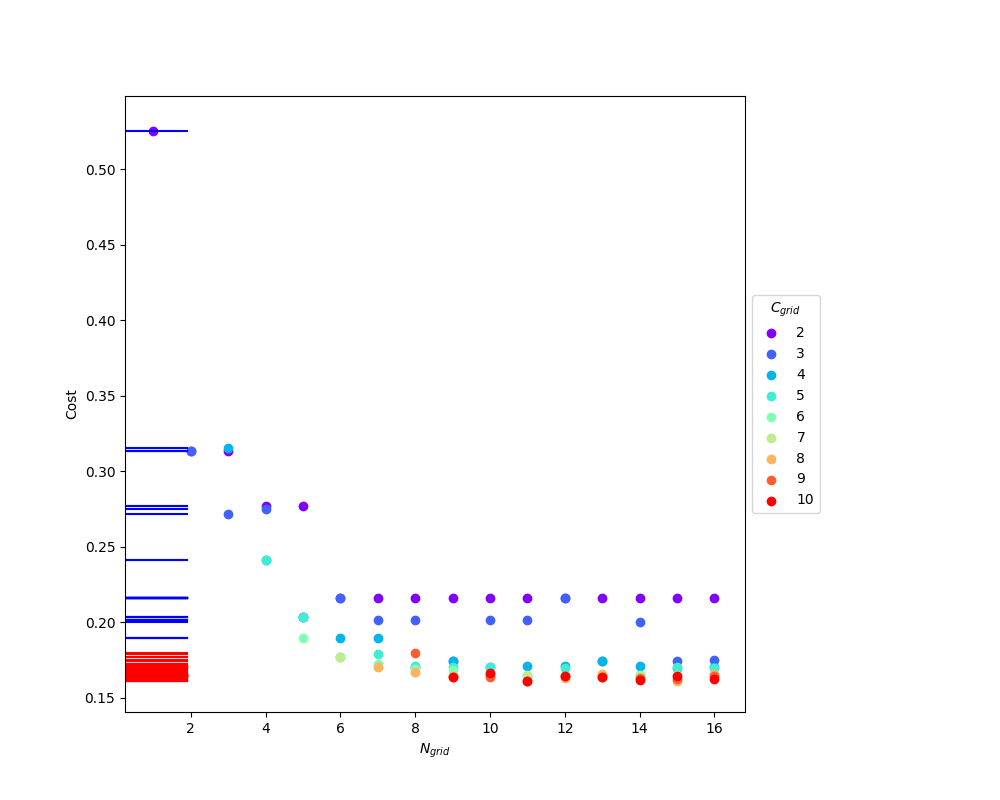}
}{%
           \caption{Grid-search criterion for WS-2-smooth\\ Most common: $N_{grid}=13$, $C_{grid}=6$}
}
\capbtabbox{%

 \begin{tabular}{||c c c c c||} 
 \hline
label &  m & n & $\rho_{w}$ & CEC \\ [0.5ex] 
 \hline\hline
 0 & 1.85 & 1.8 & 0.052 & 0 \\ 
 \hline
 1 & 2.1 & 2.0 & 0.052 & 0 \\
 \hline
 2 & 2.4 & 2.3 & 0.049 & 80 \\
 \hline
 3 & 1.9 & 2.0 & 0.051 & 0 \\
 \hline
 4 & 2.0 & 1.95 & 0.051 & 30 \\
 \hline
 5 & 2.0 & 2.5 & 0.05 & 0 \\ 
 \hline
\end{tabular}

}{%
  \caption{Parameters for WS-2-smooth}%
}
\end{floatrow}
\end{figure}

\begin{figure}[h!]
\centering
  \begin{subfigure}[b]{0.49\textwidth}
    \includegraphics[trim=4cm 2cm 2cm 2cm,clip,width=1\linewidth]{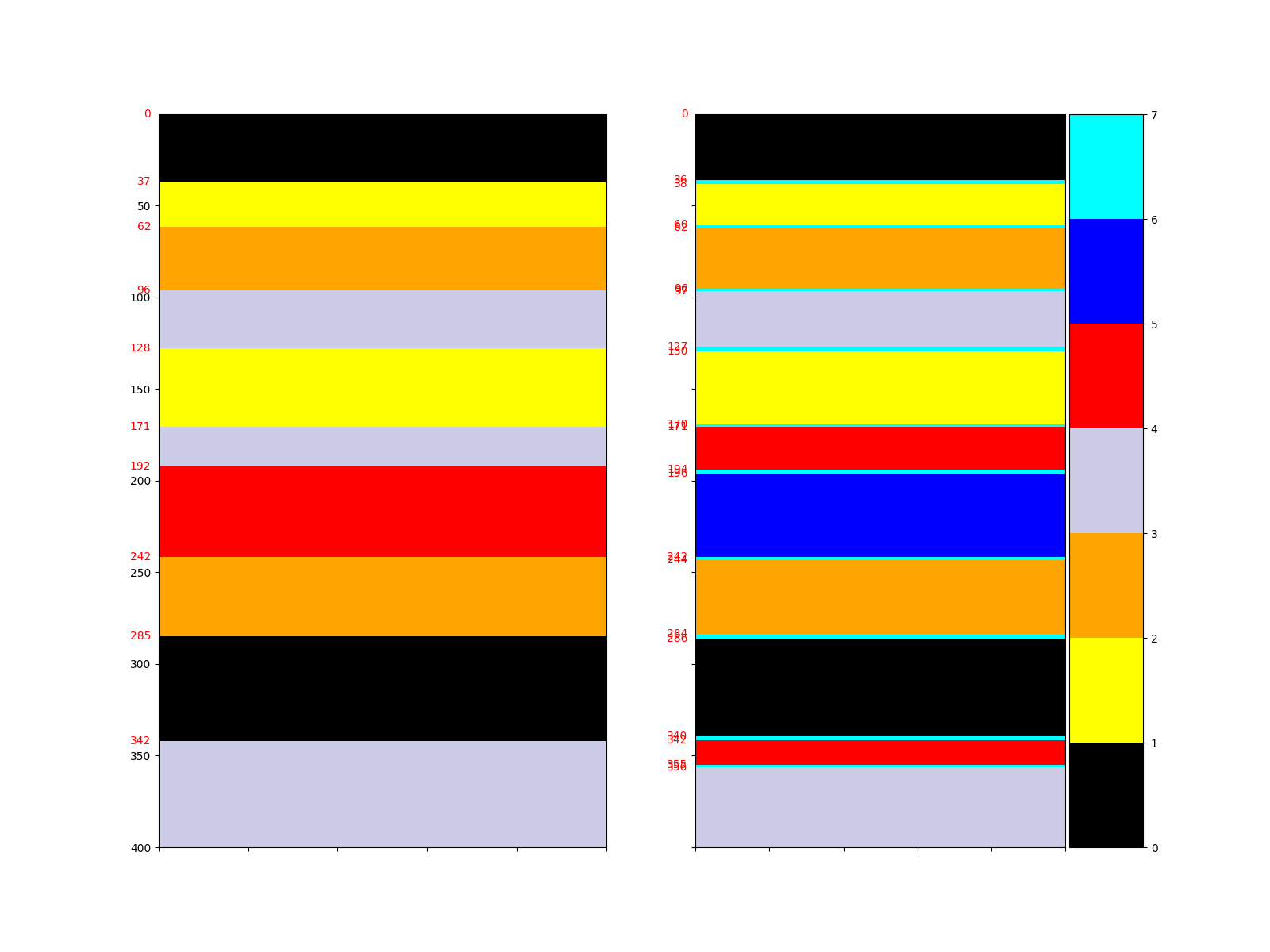}
    \caption{Most common}
  \end{subfigure}
  \begin{subfigure}[b]{0.49\textwidth}
    \includegraphics[trim=4cm 2cm 2cm 2cm,clip,width=1\linewidth]{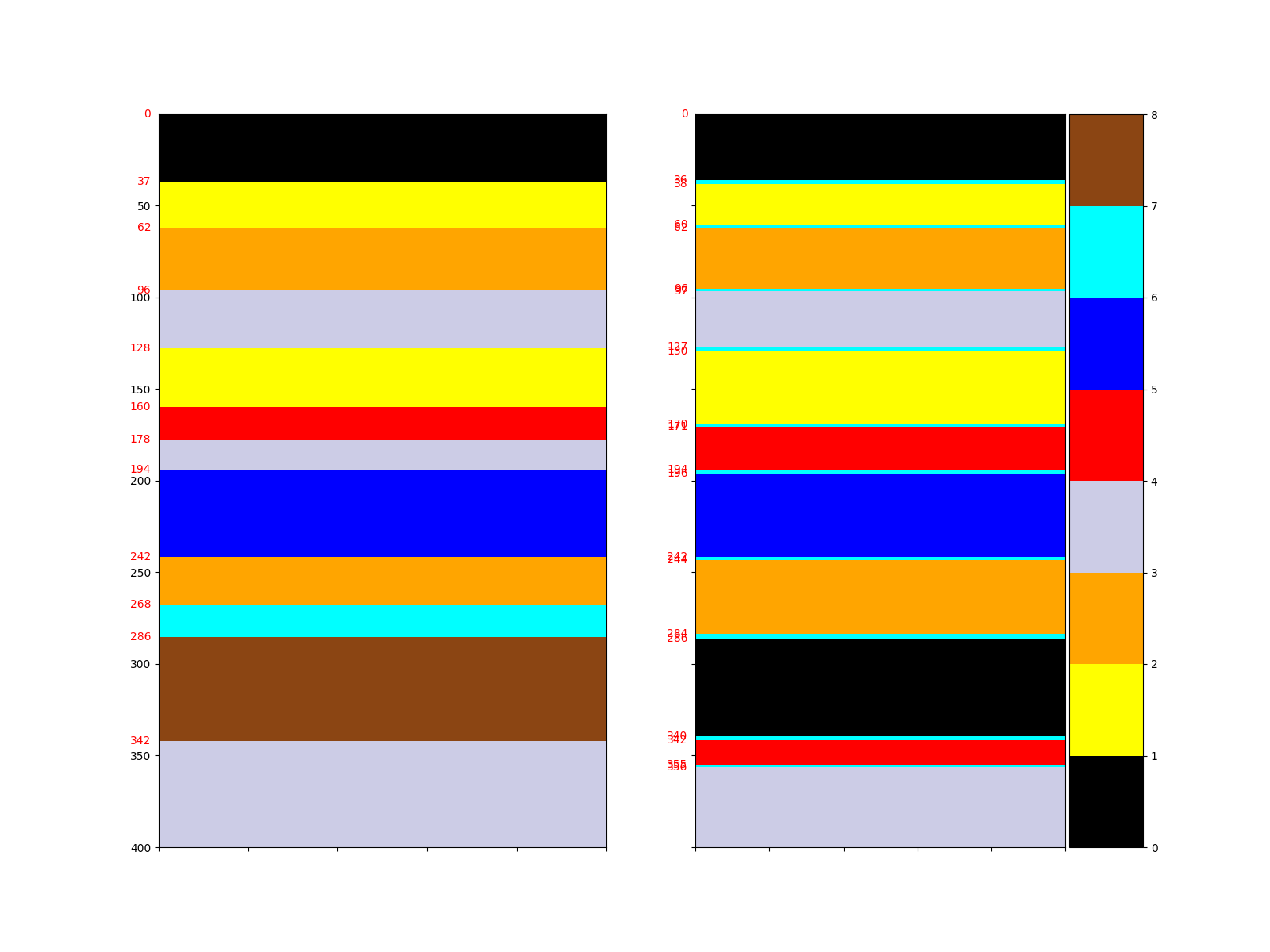}
    \caption{Lowest cost}
  \end{subfigure}
  \caption{Predictions for WS-2-smooth}
\end{figure}

\begin{figure}[h!]
\centering
  \begin{subfigure}[b]{0.49\textwidth}
    \includegraphics[trim=0cm 0cm 0cm 2cm,width=1\linewidth]{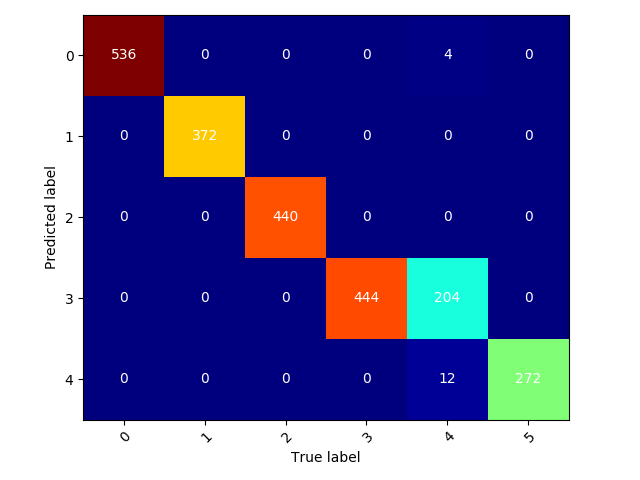}
    \caption{Most common}
  \end{subfigure}
  \begin{subfigure}[b]{0.49\textwidth}
    \includegraphics[trim=0cm 0cm 0cm 2cm,width=1\linewidth]{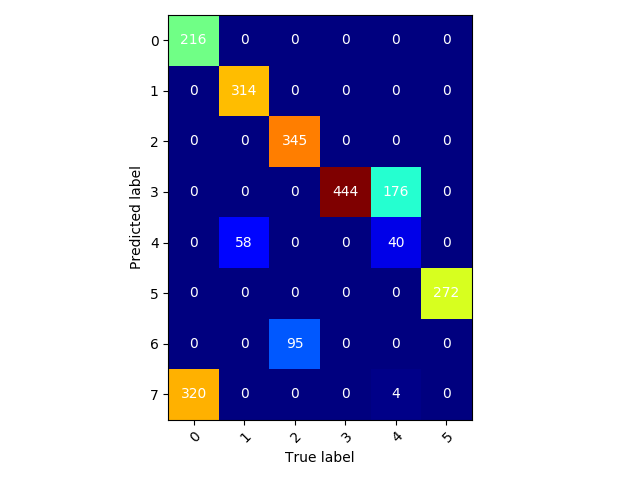}
    \caption{Lowest cost}
  \end{subfigure}
  \caption{Confusion matrices for WS-2-smooth}
\end{figure}

\end{appendices}
\end{document}